\documentclass[11pt,letterpaper]{article}

\usepackage[margin=1in]{geometry}
\usepackage{amsmath}
\usepackage{amssymb}
\usepackage{graphicx}
\usepackage{booktabs}
\usepackage{array}
\usepackage{multirow}
\usepackage{longtable}
\usepackage{adjustbox}
\usepackage{float}
\usepackage{placeins}
\usepackage{xcolor}
\usepackage{paper_style}
\usepackage{natbib}
\setcitestyle{authoryear,round,citesep={;},aysep={,},yysep={;}}
\usepackage[colorlinks=true,citecolor=PaperTeal,linkcolor=black,urlcolor=PaperNavy]{hyperref}
\usepackage{url}

\graphicspath{{figures/}}
\renewcommand{\topfraction}{0.9}
\renewcommand{\bottomfraction}{0.9}
\renewcommand{\textfraction}{0.12}
\renewcommand{\floatpagefraction}{0.75}

\title{Joules-to-GDP: How Much Economic Value Do AI Systems Produce per Unit of Energy?}
\author{Amirreza Zeinali$^{*1}$ \enspace$\cdot$\enspace Jon Saad-Falcon$^{*1}$ \enspace$\cdot$\enspace Avanika Narayan$^{*1}$\\
Mersad Abbasi$^{2}$ \enspace$\cdot$\enspace Jonathan S. Hartley$^{3}$ \enspace$\cdot$\enspace John Hennessy$^{1}$\\
Christopher Re$^{1}$ \enspace$\cdot$\enspace Azalia Mirhoseini$^{1}$\\[0.9em]
\small $^{1}$Stanford University \enspace$\cdot$\enspace $^{2}$University of California, Berkeley\\
\small $^{3}$The University of Texas at Austin}

\date{}
\hypersetup{
  pdftitle={Joules-to-GDP: How Much Economic Value Do AI Systems Produce per Unit of Energy?},
  pdfauthor={Amirreza Zeinali, Jon Saad-Falcon, Avanika Narayan, Mersad Abbasi, Jonathan S. Hartley, John Hennessy, Christopher Re, Azalia Mirhoseini}
}

\begin{document}
\frenchspacing
\raggedbottom
\color{PaperInk}
\begin{paperopening}
\renewcommand{\baselinestretch}{1}\selectfont
\setlength{\parskip}{0pt}
{\centering
  \begin{minipage}[c]{\linewidth}
    \centering
    {\fontsize{21}{24}\selectfont\bfseries\color{PaperNavy}Joules-to-GDP:\par}
    \vspace{2pt}
    {\fontsize{16}{18.5}\selectfont How Much Economic Value Do AI Systems\\
    Produce per Unit of Energy?\par}
  \end{minipage}
  \par\vspace{14pt}
  {\fontsize{11.2}{13.4}\selectfont\paperauthors\par}
}
\vspace{17pt}
{\color{PaperRule}\hrule height 0.5pt}
\vspace{9pt}
{\fontsize{11}{13}\selectfont
\setlength{\parindent}{0pt}
AI systems are increasingly used to complete economically valuable tasks in various professions, such as software engineering and finance. This motivates the question: \textit{how much economic value does an AI system produce per unit of energy?} To answer it, we introduce Joules-to-GDP, a framework that measures output dollar value minus AI execution cost per unit of energy, which we call Net Value per Joule (NVJ), and expert time saved per unit of energy, or Time Saving per Joule (TSJ). We then apply the framework to 220 GDPval tasks across 44 occupations. Four findings stand out. \textbf{First,} the most valuable work is not always the most efficient. For example, engineering work such as CAD design yields about 10\% more net value per task than software development such as building websites, yet the latter yields about 27\% more per joule. \textbf{Second,} defaulting to the most capable model can mean less total value from the same energy budget. Of the five tested models, GLM 5.2 delivers the most net value, \$72.27 per task on average, using the energy of about four hours of active laptop use, while Gemma 4 31B achieves 56\% of that with 23\% of the energy, for 2.5x the NVJ and 2.8x the TSJ. \textbf{Third,} for the same model, NVJ and TSJ vary widely across different accelerators, and adding GPUs can raise energy use for modest gains: eight H100s use 3.2x the energy of one to cut completion time by only 22\%, at similar net value. More capable models and more hardware must therefore earn their extra energy with enough gains in output value. \textbf{Finally,} the work of experts who use AI is harder for AI to match than that of experts working alone. Measured against it, GLM 5.2’s NVJ falls by 70\% and its TSJ by 65\%. Taken together, these findings justify economic return per joule as an objective for selecting professional tasks and improving language models and accelerators, and as a basis for examining AI’s effect on the economy.
\par}
\vspace{10pt}
{\color{PaperRule}\hrule height 0.5pt}
\vspace{7pt}
{\fontsize{10.3}{12.3}\selectfont
\urlstyle{same}
\noindent\begin{tabular}{@{}m{12pt}@{\hspace{6pt}}p{\dimexpr\linewidth-18pt\relax}@{}}
\href{https://github.com/HazyResearch/joules_to_gdp}{\includegraphics[width=12pt]{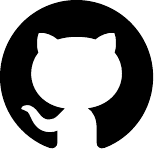}} &
\textbf{Code:} \href{https://github.com/HazyResearch/joules_to_gdp}{\color{PaperTeal}\nolinkurl{https://github.com/HazyResearch/joules_to_gdp}} \\[3pt]
\href{https://huggingface.co/datasets/ScalingIntelligence/joules-to-gdp}{\includegraphics[width=12pt]{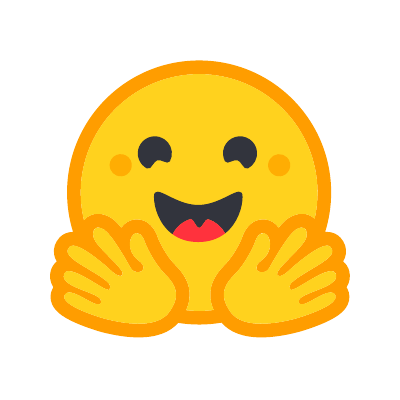}} &
\textbf{Dataset:} \href{https://huggingface.co/datasets/ScalingIntelligence/joules-to-gdp}{\color{PaperTeal}\nolinkurl{https://huggingface.co/datasets/ScalingIntelligence/joules-to-gdp}}
\end{tabular}\par}
\end{paperopening}
\begingroup
\renewcommand{\thefootnote}{\fnsymbol{footnote}}
\footnotetext[1]{Equal contribution.}
\endgroup

\section{Introduction}
\label{sec:introduction}

Existing studies assess an AI system's economic value through proxies such as output
quality and estimated worker time saving, and compare task performance with
API cost or completion time
\citep{patwardhan2025gdpval,wan2026economic,erol2025costofpass}.
Meanwhile, energy benchmarks measure how much energy AI systems consume, but
do not quantify the economic value of the professional outputs produced with
that energy
\citep[see Appendix~\ref{app:related} for related work]{chung2025mlenergy,tripathy2026swenergy}.
We connect these approaches by measuring the economic value of professional
outputs alongside the energy required to produce them.

For this study, we use energy as our resource measure for three reasons. (1) A joule is a stable physical
unit, while API providers can change prices. (2) End-to-end energy combines
model and hardware choice, tool use, waits, and retries in one measure.
(3) As we show, energy reveals differences that API cost or completion time
can miss.\par

\begin{figure}[t]
  \centering
  \begin{paperfigureframe}[0.90]
    \includegraphics[width=\linewidth]{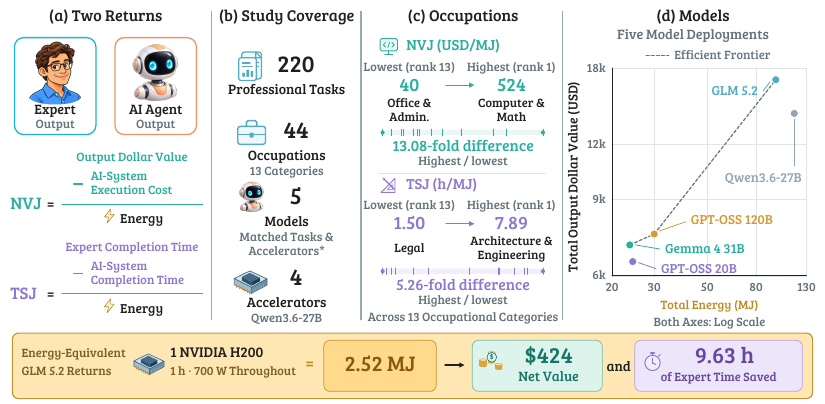}
    \end{paperfigureframe}
  \caption{\textbf{Overview of Joules-to-GDP.} (a) NVJ and TSJ express dollar and expert time benefits per joule. (b) The study compares occupations, models, and accelerators. (c) Occupation returns vary widely, with different NVJ and TSJ leaders. (d) More valuable output can require more energy; the frontier uses unweighted totals on log scales. The bottom panel uses 2.52~MJ, equivalent to one H200 drawing 700~W for an hour or about 23 hours of active laptop use.}
  \label{fig:paper-overview}
\end{figure}

Having measured the energy used to produce each AI output, we then assess its
economic value in two ways: (1) net dollar value and (2) expert time saving
from using the AI system. We express economic return using \textbf{Net Value per Joule (NVJ)} in USD/MJ and
\textbf{Time Saving per Joule (TSJ)} in hours/MJ. We use both because the work
with the greatest dollar return need not save the most expert time, as our
results show. Figure~\ref{fig:paper-overview} summarizes this evaluation.

We apply these metrics to 220 public GDPval benchmark tasks across 44
occupations and 13 categories, weighted by sector GDP and employment.
We then investigate three questions.

\textbf{Q1. How does economic return per joule vary across occupations?}
Across occupation groups, concierge and recreation work, such as preparing
travel itineraries, leads in success rate before accounting for energy, while
engineering leads in net dollar value per task. Accounting for energy shifts
the success-rate leader to investigation work, such as turning retail
investigation notes into reports, and the net-value leader to software
development, such as building website features. Factory supervision, such as
planning schedules to clear backlogs, falls from second in time saving per
task to sixth in time saving per joule. These returns depend jointly on
successful completion, the net dollar value or time saving each output
represents, and energy consumption. The expert reference used to assess AI
outputs also matters. Comparing against an expert who used AI changes the
measured returns (Section~\ref{sec:categories}).

\textbf{Q2. How do models trade off economic value and energy?}
Among five evaluated models, GLM~5.2 produces the highest average output
dollar value, at \$74.19 per task, and saves 1.64 hours of expert time using
0.430~MJ, equivalent in energy to about four hours of active laptop use. By
comparison, Gemma~4~31B produces more than half that output dollar value and
saves about two-thirds as much expert time using less than a quarter of the
energy, yielding approximately 2.5 times the NVJ and 2.8 times the TSJ.
To capture both sides of this tradeoff, we use NVJ and TSJ to compare returns
for the same amount of energy, alongside value--energy frontiers that show
the output dollar value or time saving achieved at each model's measured
energy use. Both views matter because a high return per joule can coexist
with less valuable output per task. We also find that energy reveals
differences that cost and completion time miss. GPT-OSS~120B uses 68\% less
energy than GLM~5.2 despite nearly identical execution cost and completion
time (Section~\ref{sec:systems}).

\textbf{Q3. How do accelerator type and count affect return per joule?}
Across four tested single-accelerator deployments of Qwen3.6-27B, B200 leads
NVJ and TSJ. We then vary accelerator count with the same Qwen3.6-27B model.
One and eight H100s produce nearly identical net dollar value, yet eight H100s
reduce completion time by only 22\% while using 3.24 times the energy
(Section~\ref{sec:hardware-runs}).

Our framework uses measured energy to assess an AI system's economic value
in occupational work and compare models and accelerators for deployment.
Optimizing NVJ and TSJ targets unsuccessful attempts that consume
energy and additional computation that delivers little extra value, two
shortcomings of the evaluated systems. These targets can guide improvements in
models, accelerators, software, and task routing for deployments under real
energy constraints, including expanding on-premises and edge use
\citep{deloitte2026aiinfra}. Our estimates use benchmark tasks, recorded expert
times, and occupation median wages. Whether the same returns hold beyond this
benchmark remains to be tested.

\section{Methods}
\label{sec:methods}
\begingroup

We value each AI output using an expert reference, judge the output,
measure energy, and calculate AI-system execution cost before aggregating across occupations. An
\emph{AI system} is the model, agent harness, inference backend, accelerator,
and batching policy; a \emph{run} is one execution on a task. The main analysis
uses one run per task for each configuration. We judge that output,
record its AI-system completion time, and charge its measured energy and
AI-system execution cost.
We also collected four independent runs per task to test repeated-attempt
sensitivity (Appendix~\ref{app:reference-results}, Figure~\ref{fig:thresholds}).

\subsection{Dataset and expert references}
\label{sec:dataset}

We evaluate 220 public GDPval professional tasks spanning 44 occupations
in nine U.S. economic sectors, with five tasks
per occupation \citep{patwardhan2025gdpval}. Each task contains a professional
request, source files, grading rubric, and target deliverable. Deliverables
include documents, spreadsheets, presentations, code, and media. GDPval drew
the tasks from professional work and applied multiple rounds of review;
Appendix~\ref{app:reference-protocol} gives dataset details, and
Appendix~\ref{app:contamination} reports supplementary dataset checks.

For each task, an expert reference consists of a finished deliverable
and the recorded time the expert took to produce it. We use two reference
conditions. The \textbf{\textit{expert-only reference}} was produced by an expert
without AI assistance, and the \textbf{\textit{expert-with-AI reference}} was
produced by an expert using AI. Each condition has its own deliverable and
recorded expert completion time.
We use the expert-only reference as primary because its deliverable and output
dollar value are independent of a particular assisting AI model. The
expert-with-AI reference provides a secondary comparison for assisted work.

The expert-only record covers task setup, source review, execution, and
checking without AI use. The expert-with-AI record covers the expert's
setup, prompting, monitoring, review, and correction, including waits for
AI-generated output.
AI-system completion time is elapsed time until a run stops, including losses
and timeouts; it covers inference, tool use, and task-specific setup within
the request.

Let \(G_i^b\) denote the dollar value assigned to an AI output on task \(i\)
if it wins or ties against expert reference \(b\). Let task \(i\) belong to
occupation \(o(i)\), and let
\(b\in\{\mathrm{E},\mathrm{E{+}AI}\}\) identify the expert reference.
Following the original GDPval valuation approach \citep{patwardhan2025gdpval}, we multiply expert completion time by the occupation's median hourly wage:
\begin{equation}
    G_i^b = H_i^b W_{o(i)}^{\mathrm{median}},
    \label{eq:task-value}
\end{equation}
where \(H_i^b\) is the recorded expert completion time for the selected reference. The term \(W_{o(i)}^{\mathrm{median}}\) is the
occupation's median hourly wage from the annual Occupational Employment and
Wage Statistics (OEWS) program, the U.S. national statistical reference for
occupational employment and wage estimates. Appendix~\ref{app:weights}
documents the wage source and GDP-based aggregation.

\subsection{Blinded evaluation}
\label{sec:judge}

The AI system receives the task and source files but not its rubric.
We use the multimodal Gemini 3.1 Flash-Lite as the judge throughout the main
analysis \citep{google2026gemini31flashlite}. It compares the AI-system
deliverable with the selected expert reference in a blinded pairwise comparison.
The judge receives the task, rubric,
source evidence, and identity-masked candidates. A shared pipeline extracts
sanitized semantic content and rendered visual evidence, enabling the judge to
assess the documents, spreadsheets, presentations, code, and media represented
in GDPval. Candidate positions are balanced. The judge chooses either candidate
or a tie. Let \(Y_i^b=1\) when the AI
system wins or ties and \(Y_i^b=0\) when it loses. Following GDPval's blinded
win-rate approach \citep{patwardhan2025gdpval}, we measure task success by the
win--tie rate: the share of AI outputs judged a win or tie against the expert
reference. The output dollar value
contribution is \(Y_i^bG_i^b\), which equals \(G_i^b\) for a win or tie and zero for a
loss. We choose this binary win-or-tie/loss rule to be conservative about partial credit.
A loss contributes zero even if the output has some utility, while a tie receives
the full output dollar value.
In a judge-consistency check, Gemini 3.1 Flash-Lite, Claude Haiku~4.5,
and GPT-5 mini showed 72.3--77.2\% pairwise exact agreement. Gemini's
win--tie/loss decisions also agreed with human experts on 23 of 32 GLM~5.2
comparisons (71.9\%).
Appendix~\ref{app:judge} details the evidence pipeline, candidate balancing,
and judge settings; Appendix~\ref{app:judge-consistency} reports agreement
among the three judges and with human reviewers.
Figure~\ref{fig:methods-pipeline}
illustrates the accounting for a GDPval task.

\begin{figure}[t]
    \centering
    \begin{paperfigureframe}[0.90]
    \includegraphics[width=\linewidth]{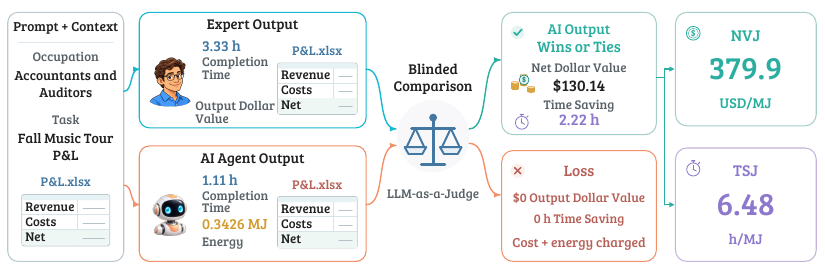}
    \end{paperfigureframe}
    \caption{\textbf{An illustrative GDPval example shows the evaluation workflow.} An Accountants and Auditors task, Fall Music Tour P\&L, compares the expert-only output with the AI-agent output through a blinded LLM-as-a-Judge comparison. A win or tie earns output dollar value and time saving; every attempt incurs AI-system execution cost and energy.}
    \label{fig:methods-pipeline}
\end{figure}

\subsection{End-to-end energy accounting}
\label{sec:energy}

We measure energy with Intelligence per Watt (IPW), from immediately before
each batch's runs launch until the final run in that batch completes
\citep{saadfalcon2025ipw}.
Here, \emph{end-to-end} denotes the full execution window, including inference,
tool use, and waits. We sum the component-energy readings recorded by IPW
(e.g., GPU, CPU, and DRAM), using counter differences or integrated sampled power.
Appendix~\ref{app:execution} details the measurement procedure.

We batch concurrent requests because deployed inference commonly serves
multiple requests at the same time. The batch policy therefore represents a
realistic serving condition rather than serial execution of isolated requests.

For a batch of \(K_b\) requests, active-request attribution divides each
moment's energy among the requests still running. For request \(i\),
\begin{equation}
E_i=\int_{i\ \mathrm{running}}\frac{P_{\mathrm{batch}}(t)}{N_{\mathrm{running}}(t)}\,dt
+\frac{E_{b,0}}{K_b}.
\label{eq:active-energy}
\end{equation}
Here \(P_{\mathrm{batch}}(t)\) is batch power, \(N_{\mathrm{running}}(t)\) counts
active requests, and \(E_{b,0}\) is shared energy with no active request.
The active portion assigns energy according to how long each request actually
runs, giving longer-running requests a larger share when power and concurrency
are comparable.
When only a batch total is retained, average power is total joules divided
by batch completion time. Every request incurs energy, including losses and timeouts.
Appendix~\ref{app:energy-attribution} tests attribution sensitivity;
Figure~\ref{fig:completion-energy} shows energy use as requests finish.

\subsection{Net Value per Joule and Time Saving per Joule}
\label{sec:efficiency-metrics}

For an AI output on task \(i\), let \(C_i\), \(E_i\), and \(L_i\) be its
AI-system execution cost, measured energy, and AI-system completion time.
Its credited output dollar value is \(Y_i^bG_i^b\), where \(G_i^b\) is based on
the selected reference expert's completion time \(H_i^b\) (Equation~\ref{eq:task-value}).
The output's net dollar value is \(Y_i^bG_i^b-C_i\), and its time saving
is \(Y_i^b(H_i^b-L_i)\). For occupation \(o\), let
\(G_o^b=\sum_{i:o(i)=o}Y_i^bG_i^b\), \(C_o=\sum_{i:o(i)=o}C_i\), and
\(E_o=\sum_{i:o(i)=o}E_i\). We define
\textbf{\textit{Net Value per Joule (NVJ)}} (in USD/MJ) and
\textbf{\textit{Time Saving per Joule (TSJ)}} (in hours/MJ):
\begin{equation}
\mathrm{NVJ}_{o}^b=\frac{G_{o}^b-C_{o}}{E_{o}},
\qquad
\mathrm{TSJ}_{o}^b=
\frac{\sum_{i:o(i)=o}Y_i^b\left(H_i^b-L_i\right)}{E_{o}}.
\label{eq:occupation-efficiency}
\end{equation}
Only wins or ties contribute output dollar value or time saving; every run incurs
cost, so a loss has negative net dollar value. We report energy in MJ
(\(1\,\mathrm{MJ}=10^6\,\mathrm{J}\)).

We calculate AI-system execution cost from recorded resource
allocations and published GPU, CPU, and RAM prices. Shared node
time uses active-request shares within the recorded task-execution
window. Shared idle time inside that window is assigned once. Storage and
transfer charges are added where applicable. This method values recorded
resource use at published rates; rate sources and allocation rules are in
Appendix~\ref{app:rental-cost}.

\subsection{Two levels of GDP-based aggregation}
\label{sec:gdp-aggregation}

For each evaluated occupation, we average output dollar value, AI-system execution
cost, time saving, and energy over its observed tasks. GDP-based occupation
weights \(w_o^{\mathrm{GDP}}\) combine sector GDP shares with within-sector
employment shares. Wages do not enter these weights because the occupation's median
wage already enters output dollar value, and wage-bill weights would give
high-wage occupations additional influence. These weights define a
GDP-weighted benchmark portfolio (Appendix~\ref{app:weights}, Figure~\ref{fig:economic-weights}).
The O*NET-SOC taxonomy groups the 44 occupations into 13 categories. We
aggregate at two levels.

\noindent\textbf{Occupation-category level.} We renormalize occupation weights
within each category to sum to one, yielding category-normalized averages
for comparisons among categories.
\par

\noindent\textbf{Benchmark level.} We weight category outcomes by their category
weights, yielding GDP-weighted task averages for comparisons among AI systems.
\par

\noindent Writing \(n_o\) for the number of evaluated tasks in occupation \(o\),
\(S_o^b=\sum_{i:o(i)=o}Y_i^b(H_i^b-L_i)\),
\(W_c=\sum_{o\in c}w_o^{\mathrm{GDP}}\), and \(X\in\{G,C,S,E\}\):
\begin{equation}
\widetilde{X}_c
=\sum_{o\in c}\frac{w_o^{\mathrm{GDP}}}{W_c}\frac{X_o}{n_o},
\qquad
\widetilde{X}^{\mathrm{GDP}}=\sum_c W_c\widetilde{X}_c.
\label{eq:two-level-aggregation}
\end{equation}
We aggregate the components first and then form the NVJ and TSJ ratios.
\par
\endgroup

\section{Experiments and Results}
\label{sec:experiments}

The results follow Q1--Q3. Q1 holds the AI system fixed and varies the occupation
and expert reference. Q2 compares models on the same tasks and evaluation protocol
across value, AI-system execution cost, AI-system completion time, and energy.
Q3 varies accelerator type and count for the same model. This
separates differences in the work from differences in the AI system.

\begin{figure}[t]
    \centering
    \begin{paperfigureframe}[0.95]
    \includegraphics[width=\linewidth]{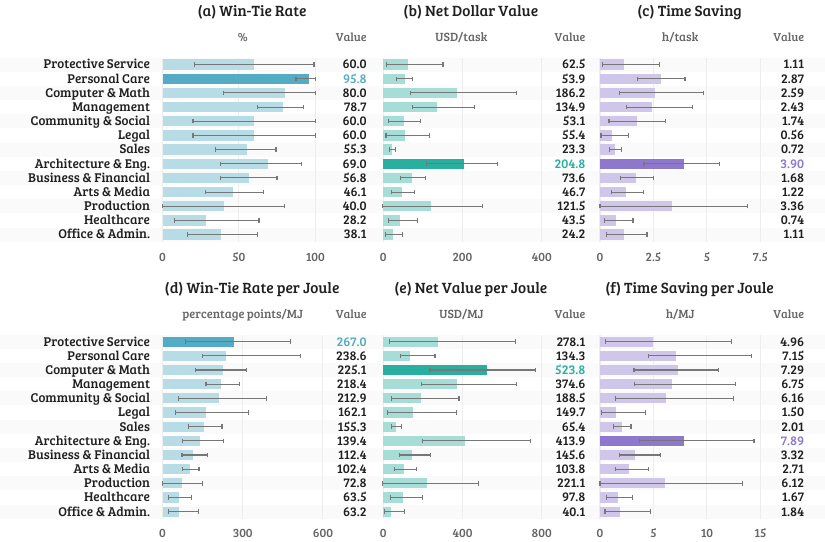}
    \end{paperfigureframe}
    \caption{\textbf{Energy changes occupation rankings, and each return metric favors different work.} GLM~5.2 against the expert-only reference, with category-normalized GDP weights. Top panels show win--tie rate, net dollar value, and time saving per task; bottom panels normalize each by energy. Categories follow the win--tie-per-MJ order; whiskers show 95\% task-composition intervals.}
    \label{fig:category-rankings}
\end{figure}

\subsection{How does economic return per joule vary across occupations? (Q1)}
\label{sec:primary-result}
\label{sec:categories}
\label{sec:thresholds}

We evaluate GLM~5.2 with OpenHands on eight NVIDIA H200 GPUs in 20-request
batches across the evaluated tasks.

\textbf{Energy changes occupation rankings, and each return metric captures a different benefit.}
Personal Care \& Service leads in win--tie rate, but once energy is included,
Protective Service leads per MJ. Net dollar value shows a similar reversal.
Architecture \& Engineering (for example, CAD design) delivers about
10\% more per task than Computer \& Mathematical (for
example, website design). However, after energy is included, Computer \&
Mathematical delivers about 27\% higher NVJ. Likewise, Production falls from second in
time saving per task to sixth in TSJ (Figure~\ref{fig:category-rankings}).

The three energy-normalized metrics also capture different benefits.
For example, Computer \& Mathematical delivers nearly four times Personal Care \&
Service's NVJ despite a lower win--tie rate per MJ, because each successful
output represents more dollar value. Time saving reveals another reversal.
Production has about 20\% lower NVJ than Protective Service, yet 23\% higher
TSJ because each success saves more expert time. High NVJ and TSJ thus combine success
with greater net dollar value or time saving relative to energy consumed
across all attempts.
Appendices~\ref{app:category-results} and~\ref{app:success-latency} give category
values and explain these rankings; Appendix~\ref{app:success-latency-categories}
lists full results, Appendix~\ref{app:category-omission} tests rank stability,
and Appendix~\ref{app:judge-consistency} checks judge agreement.

\begin{figure}[t]
    \centering
    \begin{paperfigureframe}[0.85]
    \includegraphics[width=\linewidth]{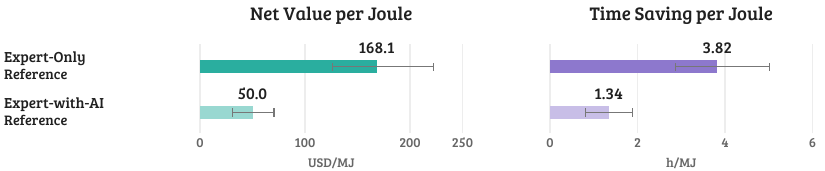}
    \end{paperfigureframe}
    \caption{\textbf{An expert-with-AI reference reduces GLM 5.2's NVJ by 70\% and TSJ by 65\%.} Only the expert deliverable and its completion time change; AI runs, costs, energy, and GDP weights stay fixed. Whiskers show 95\% task-composition intervals (Appendix~\ref{app:reference-decomposition}).}
    \label{fig:baseline-comparison}
\end{figure}

\textbf{The expert-with-AI reference is harder to beat and closer to how many professionals work.}
The expert-only reference provides a comparison independent of an assisting
model. By contrast, the expert-with-AI reference reflects the model and tools
the expert used, while testing the additional benefit of automation when experts
already use AI. Under our weighting, experts using AI spend about 23\% less
time on average, and GLM~5.2 matches or exceeds their deliverables about half
as often as the expert-only deliverables. Automation must therefore meet a
higher quality bar while replacing less expert time, lowering NVJ and TSJ
(Figure~\ref{fig:baseline-comparison}). Yet substantial time-saving potential
remains. On tasks where GLM~5.2 matches or exceeds the assisted expert's
deliverable, its completion time is about 84\% shorter on average than the assisted
expert's (Appendix~\ref{sec:passk}, Table~\ref{tab:time-decomposition}).
Figure~\ref{fig:three-way} compares time saving from expert assistance and
automation across success thresholds.

\begin{figure}[t]
    \centering
    \begin{paperfigureframe}[0.90]
    \includegraphics[width=\linewidth]{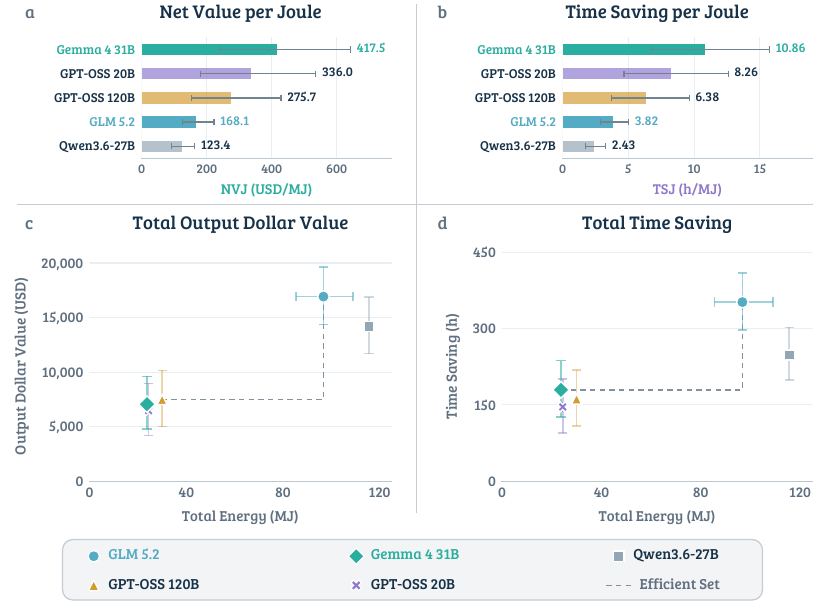}
    \end{paperfigureframe}
    \caption{\textbf{Higher total economic benefit can come with lower return per joule.} For example, GLM~5.2 delivers more total value and time saving than Gemma~4~31B, but has lower NVJ and TSJ. The top row compares GDP-weighted NVJ (a) and TSJ (b); the bottom shows unweighted total output dollar value (c) and time saving (d) versus total energy. Dashed steps connect models on each point-estimate efficient frontier. Whiskers show 95\% task-composition intervals.}
    \label{fig:pareto}
\end{figure}

\subsection{How do models trade off economic value and energy? (Q2)}
\label{sec:systems}

We serve five models locally so that we can measure end-to-end energy directly:
GLM~5.2\footnote{GLM~5.2 uses eight H200 GPUs because its FP8 serving setup
did not fit on the tested eight-H100 node used for the other four models.},
Qwen3.6-27B, Gemma~4~31B, GPT-OSS~120B, and GPT-OSS~20B
\citep{glm2026license,qwen2026license,gemma2026license,gptoss2025license}.
We swap the model while holding the evaluated tasks, expert-only reference, judge,
OpenHands harness, and vLLM inference engine fixed; the four non-GLM models
also use the same eight-H100 hardware.
Appendix~\ref{app:systems} summarizes model outcomes (Figure~\ref{fig:global-metrics});
Appendix~\ref{app:energy-attribution} tests energy attribution for the
GLM~5.2 occupation comparisons.

\textbf{Higher output need not justify its energy cost.}
GLM~5.2 produces the greatest output dollar value, \$74.19 per task, and saves
1.64 expert hours per task on average. Gemma~4~31B produces
44\% less output dollar value but uses 77\% less energy,
giving it about 2.5 times GLM~5.2's NVJ and the highest TSJ among these models.
For scale, GLM~5.2's 0.430~MJ per task is equivalent in energy to about
four hours of active laptop use\footnote{Laptop-use equivalents assume
30~W, the midpoint of the 20--40~W active-laptop range reported by
\citet{nrel2011plugload}.} or 10.2 minutes of one H200 drawing 700~W
continuously \citep{nvidiaH200}.
Qwen3.6-27B produces the second-highest output dollar value but uses the most
energy among the five models. Figure~\ref{fig:pareto} ranks the five models by GDP-weighted
NVJ and TSJ and shows unweighted total output dollar value and total time saving against
total energy across evaluated tasks. On both unweighted point-estimate efficient
frontiers, Gemma~4~31B anchors the low-energy end and GLM~5.2 the high-benefit
end. GPT-OSS~120B lies between them on the dollar-value frontier. These
tradeoffs explain why the rankings and frontiers answer complementary questions.
NVJ and TSJ measure net dollar return and expert time saved per megajoule,
while the frontiers show how much total value or time saving each model delivers
for a given energy budget. A high return per joule
can still come from a model that delivers too little total work. These
views explain why Gemma~4~31B leads in return per joule while GLM~5.2 remains
on the frontier by delivering more value at higher energy use.
Occupation-category rankings are broadly similar across models
(Appendix~\ref{app:systems}).

\begin{table}[t]
\centering
\caption{\textbf{GLM~5.2 and GPT-OSS~120B have similar cost and completion time but different energy returns.}
AI-system execution cost, AI-system completion time, and energy are GDP-weighted
task averages; the three efficiency
ratios share each system's net-dollar-value numerator. Colors identify systems.
Bars show point estimates on a separate zero-to-maximum scale within each
efficiency column; longer bars mean greater net value per unit of resource.}
\label{tab:system-cost-time-energy}
\begingroup
\small
\setlength{\tabcolsep}{3pt}
\renewcommand{\arraystretch}{1.18}
\definecolor{ModelGLM}{HTML}{264B73}
\definecolor{ModelQwen}{HTML}{5C809D}
\definecolor{ModelGemma}{HTML}{16817A}
\definecolor{ModelOSS120}{HTML}{B58C4D}
\definecolor{ModelOSS20}{HTML}{8A7095}
\newcommand{\modelkey}[2]{\textcolor{#1}{\rule{4pt}{4pt}}\hspace{3pt}#2}
\newcommand{\metricbar}[3]{%
  \makebox[28pt][l]{\rlap{\textcolor{black!9}{\rule{28pt}{3.5pt}}}%
  \textcolor{#1}{\rule{#2pt}{3.5pt}}}\hspace{2pt}\makebox[25pt][r]{#3}}
\begin{tabular*}{\linewidth}{@{\extracolsep{\fill}}lrrrccc@{}}
\toprule
 & \multicolumn{3}{c}{GDP-weighted task averages} & \multicolumn{3}{c}{Net value per} \\
\cmidrule(lr){2-4}\cmidrule(lr){5-7}
Model & \shortstack{Execution\\cost (USD)} & \shortstack{Completion\\time (min)} & \shortstack{Energy\\(MJ)} & \shortstack{dollar of AI cost\\(USD/USD)} & \shortstack{AI hour\\(USD/hour)} & \shortstack{megajoule\\(USD/MJ)} \\
\midrule
\modelkey{ModelGLM}{GLM 5.2} & 1.92 & 22.7 & 0.430 & \metricbar{ModelGLM}{22.276}{37.69} & \metricbar{ModelGLM}{16.898}{191.3} & \metricbar{ModelGLM}{11.275}{168.1} \\
\modelkey{ModelQwen}{Qwen3.6-27B} & 4.12 & 51.0 & 0.516 & \metricbar{ModelQwen}{9.137}{15.46} & \metricbar{ModelQwen}{6.626}{75.0} & \metricbar{ModelQwen}{8.276}{123.4} \\
\modelkey{ModelGemma}{Gemma 4 31B} & 0.86 & 7.7 & 0.097 & \metricbar{ModelGemma}{28.000}{47.38} & \metricbar{ModelGemma}{28.000}{316.9} & \metricbar{ModelGemma}{28.000}{417.5} \\
\modelkey{ModelOSS120}{GPT-OSS 120B} & 1.98 & 22.2 & 0.137 & \metricbar{ModelOSS120}{11.283}{19.09} & \metricbar{ModelOSS120}{9.018}{102.1} & \metricbar{ModelOSS120}{18.489}{275.7} \\
\modelkey{ModelOSS20}{GPT-OSS 20B} & 1.12 & 15.3 & 0.103 & \metricbar{ModelOSS20}{18.339}{31.03} & \metricbar{ModelOSS20}{12.040}{136.3} & \metricbar{ModelOSS20}{22.537}{336.0} \\
\bottomrule
\end{tabular*}
\endgroup
\end{table}

\subsubsection{Net value per dollar of AI cost and per AI hour}
\label{sec:multiplier}

To test whether energy as a denominator reveals a dimension of efficiency
that monetary cost and AI-system completion time alone miss, we define two comparison metrics:
\begin{equation}
\text{Net value/dollar of AI cost}=\frac{G-C}{C},\qquad
\text{Net value/AI hour}=\frac{G-C}{L}.
\label{eq:net-value-per-dollar}
\end{equation}
Here \(G\), \(C\), and \(L\) are GDP-weighted averages of credited output dollar
value, AI-system execution cost, and AI-system completion time. Costs and durations include
losses and timeouts; concurrent runs each contribute their AI-system completion time to \(L\), so an
AI hour is a request-hour. Table~\ref{tab:system-cost-time-energy} compares
resource use and the three efficiency ratios, each with numerator \(G-C\).
Appendix~\ref{app:rental-cost} details resource prices and cost allocation;
Appendix~\ref{app:provider-cost-sensitivity} tests alternative provider rates.

\textbf{Energy reveals differences beyond AI-system execution cost and completion time.}
GLM~5.2 and GPT-OSS~120B have nearly the same GDP-weighted average AI-system execution cost and
AI-system completion time (GPT-OSS~120B costs about 3.1\% more and is about 2.1\% faster), yet GPT-OSS~120B uses
68\% less energy. Despite its lower net value, GPT-OSS~120B delivers more per
megajoule. By contrast, GLM~5.2 delivers more per dollar of AI cost and per
AI hour. This reversal
is possible because execution cost reflects allocated resources
and elapsed time, and completion time records those elapsed minutes; measured
energy instead integrates GPU, CPU, and DRAM power during computation, memory
access, and waiting \citep{chung2025mlenergy}. Similar costs and durations can
therefore mask different joule totals.

\subsection{How do accelerator type and count affect return per joule? (Q3)}
\label{sec:hardware-runs}

Qwen3.6-27B is used for the accelerator experiments because it fits on a single
accelerator, allowing both one-accelerator sweeps and comparison with the
one-node, eight-H100 configuration above.

\begin{figure}[t]
    \centering
    \begin{paperfigureframe}[0.90]
    \includegraphics[width=\linewidth]{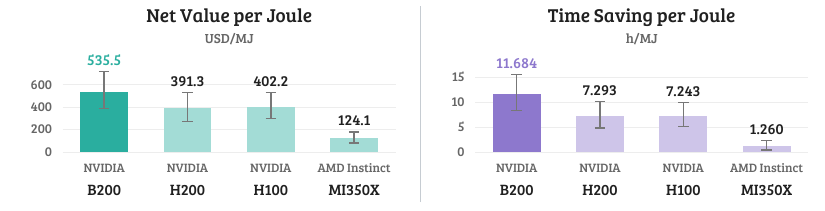}
    \end{paperfigureframe}
    \caption{\textbf{Accelerator choice changes return per joule.} Qwen3.6-27B, one accelerator, expert-only reference, and GDP weights. Whiskers show 95\% task-composition intervals.}
    \label{fig:hardware-medians}
\end{figure}

\textbf{Accelerator type materially
changes the observed return per joule.}
We run Qwen3.6-27B with OpenHands on one NVIDIA B200, H200, or H100 and one AMD
Instinct MI350X, using the same expert-only reference and judge. B200 delivers
the highest GDP-weighted NVJ and TSJ among these deployments
(Figure~\ref{fig:hardware-medians}). With Qwen3.6-27B held fixed, differences
in accelerator architecture, memory bandwidth, and power draw help explain
differences in inference time and energy use, and hence the spread in NVJ and
TSJ across devices.

\begin{figure}[t]
    \centering
    \begin{paperfigureframe}[0.90]
    \includegraphics[width=\linewidth]{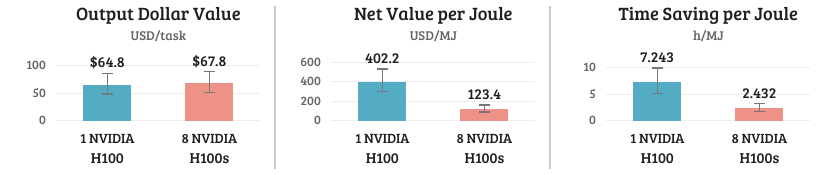}
    \end{paperfigureframe}
    \caption{\textbf{Eight H100s deliver similar value but lower return per joule than one.} Qwen3.6-27B, expert-only reference, and GDP weights. Whiskers show 95\% task-composition intervals. Appendix~\ref{app:hardware} gives the full comparison.}
    \label{fig:qwen-h100-count}
\end{figure}

\textbf{One H100 delivers more net value per joule, whereas eight deliver more per AI hour.}
Across all tasks, moving Qwen3.6-27B from one to eight H100s reduces
GDP-weighted mean AI-system completion time by 22\% but uses 3.24 times the
energy while delivering nearly the same output dollar value. Eight H100s
yield more net value per AI hour, while one H100 yields higher NVJ and TSJ
(Figure~\ref{fig:qwen-h100-count}).

Tensor parallelism splits inference across eight H100s, helping reduce
completion time, but GPU communication and synchronization, together with
sequential tool calls, limit the end-to-end gain \citep{chung2025mlenergy}.
All eight GPUs draw power throughout the run, so the shorter duration does
not offset their higher aggregate power use. With output value nearly
unchanged, this explains why eight H100s improve return per AI hour while
one H100 delivers higher NVJ and TSJ.

\section{Limitations and Future Work}
\label{sec:discussion}

\noindent\textbf{Benchmark scope and evaluation.}
Following GDPval, output dollar value uses expert completion time and occupation
median wage, not measured GDP. GDP-contributing occupations and GDP-based weights
motivate the framing; future work could validate the proxy with independent
expert valuations and test broader task samples.

\noindent\textbf{System coverage.}
Controlled accelerator sweeps vary hardware type and count. Future comparisons
could vary model, harness, backend, precision, batching, and retries individually
to isolate their effects and test portability beyond OpenHands.

\noindent\textbf{Measurement and deployment.}
Energy covers metered GPU, CPU, and DRAM; execution cost uses published rates
(Appendix~\ref{app:rental-cost}). Deployment studies could meter facility
overhead and estimate expert review, coordination, liability, and adoption costs.

\noindent\textbf{Toward optimization.}
Future work can test NVJ and TSJ as training signals and use them for routing
and stopping. Such policies should spend computation only when expected value
or time saving warrants the energy, subject to quality and reliability.

\section{Conclusion}
\label{sec:conclusion}

Joules-to-GDP connects judged professional work with the energy used to
produce it. The evaluated tasks and systems illustrate a framework rather than
a comprehensive survey. In deployments with limited energy, this accounting
makes computation allocation an explicit choice across tasks, models, and
accelerators. Comparing total benefit with return per joule can then guide
routing, stopping, and serving decisions toward higher economic value and
greater expert time savings within that budget.

\FloatBarrier
\label{end:main}
\section*{Reproducibility Statement}

We will release our task-level evaluation dataset with expert annotations,
blinded model judgments, completion times, execution costs, and energy
measurements. The GDPval task dataset is public, and the five evaluated models
have publicly available weights. The release will identify the model and judge
versions and give access details for any restricted reference materials.

We will also release the study-authored harness configuration,
energy-accounting and aggregation code, figure and table builders, fixed
analysis seeds, and a source inventory with version identifiers and hashes.
These materials support regeneration of reported results from the same recorded
inputs. The methods and appendix document expert references, judging, run
selection, cost boundaries, measured energy components, economic weights, and
validation checks. End-to-end reruns require compatible hardware and serving
software and may produce different outputs or energy measurements.

\section*{Acknowledgements}
We thank the Scaling Intelligence Lab and Hazy Lab for constructive feedback on this paper. We are especially grateful to Snorkel AI for its feedback and for providing the materials and input that made this research possible. At Snorkel AI, we thank Ramya Ramakrishnan, Vincent Sunn Chen, Srikar Kodati, Amanda Dsouza, and Paroma Varma. We thank Yasa Baig, Rishi Bommasani, Catherine Dong, Yuzhen Mao, Shayan Talaei, and Thoughtful Lab for their feedback. We thank Modal, the Laude Institute, Google Cloud Platform, IBM Research, and Stanford HAI for providing compute resources that enabled our experiments. We also gratefully acknowledge the support of NSF under Nos. CCF2247015 (Hardware-Aware) and 2522743 (AIMing); DARPA under No. HR00112520038 (Fallingwater); NSTXL under No. N00164-23-9-G057-01 (Energy-Efficient AI Hardware); ONR under No. N000142312633 (Deep Signal Processing); Stanford HAI under No. 345077 (Evo); Google DeepMind; Google Research; Google Cloud; NXP, Xilinx, LETI-CEA, Intel, IBM (Code Generation), Microsoft, NEC, Toshiba, TSMC, ARM, Hitachi, BASF, Accenture, Ericsson, Qualcomm, Analog Devices, Salesforce, Total, Lightspeed Venture Partners, the Laude Institute, Prime Intellect, Anthropic, the HAI-GCP Cloud Credits for Research program, the Stanford Data Science Initiative (SDSI), Knight-Hennessy Scholarship, JP Morgan AI/ML Fellowship, Stanford EDGE Fellowship, GEM Fellowship, and members of the Stanford DAWN project: Meta, Google, and VMWare; and members of the Stanford SEAMS project: IBM and Felicis Ventures. We also gratefully acknowledge computing resources provided by Stanford's Marlowe Computing Instrument. The U.S. Government is authorized to reproduce and distribute reprints for Governmental purposes notwithstanding any copyright notation thereon. Any opinions, findings, and conclusions or recommendations expressed in this material are those of the authors and do not necessarily reflect the views, policies, or endorsements, either expressed or implied, of NSF, ONR, DARPA, or the U.S. Government.

\bibliography{references}
\bibliographystyle{references}

\FloatBarrier
\appendix
\begingroup
\setcounter{topnumber}{6}
\setcounter{bottomnumber}{4}
\setcounter{totalnumber}{10}
\renewcommand{\topfraction}{0.95}
\renewcommand{\bottomfraction}{0.9}
\renewcommand{\textfraction}{0.05}
\renewcommand{\floatpagefraction}{0.85}
\setlength{\floatsep}{10pt plus 2pt minus 2pt}
\setlength{\textfloatsep}{12pt plus 2pt minus 2pt}
\setlength{\intextsep}{10pt plus 2pt minus 2pt}
\setlength{\LTpre}{6pt}
\setlength{\LTpost}{6pt}
\captionsetup[table]{skip=5pt}
\linespread{1}\selectfont
\setlength{\parskip}{2pt plus 0.5pt minus 0.5pt}
\clubpenalty=10000
\widowpenalty=10000
\displaywidowpenalty=10000
\makeatletter
\setlength{\@fpsep}{10pt plus 2pt minus 2pt}
\renewcommand{\section}{\@startsection{section}{1}{\z@}
  {-2.5ex \@plus -0.5ex \@minus -0.2ex}{1ex \@plus 0.2ex}
  {\normalfont\Large\bfseries}}
\renewcommand{\subsection}{\@startsection{subsection}{2}{\z@}
  {-2ex \@plus -0.5ex \@minus -0.2ex}{0.8ex \@plus 0.2ex}
  {\normalfont\large\bfseries}}
\renewcommand{\paragraph}{\@startsection{paragraph}{4}{\z@}
  {1.75ex \@plus 0.5ex \@minus 0.2ex}{-1em}
  {\normalfont\normalsize\bfseries}}
\let\preprintsection\section
\makeatother

\section{Related Work}
\label{app:related}

\textbf{Professional work and economic value.} \emph{GDPval}
\citep{patwardhan2025gdpval} evaluates deliverables;
\emph{APEX-Agents} and the \emph{Remote Labor Index}
\citep{vidgen2026apexagents,mazeika2025rli} compare AI systems with experts
on longer tasks. Related benchmarks extend this approach to research, IT, web,
desktop, organizational settings, and economic simulations
\citep{starace2025paperbench,jha2025itbench,xie2024osworld,
backlund2025vending}. Controlled and field studies complement these benchmarks
by measuring changes in workers' completion time, output quality, and productivity
\citep{noy2023experimental,brynjolfsson2025generative}.

\textbf{Occupations and the wider economy.} \emph{GPTs Are GPTs}
\citep{eloundou2023gpts} estimates which occupational tasks language models
could affect. The \emph{Anthropic Economic Index}
\citep{tamkin2025tasks,tamkin2025productivity} uses real Claude conversations
to study where AI is being used at work and estimate possible time saving,
while \emph{Economic Evaluations of Language Models} \citep{wan2026economic} tests
tasks across the U.S. occupational economy and estimates how much work time current
models could save. Related studies connect AI capabilities, use, and adoption
to workers, occupations, firms, industries, and the broader economy
\citep{felten2021occupational,shao2025future,bonney2026microstructure,
acemoglu2024simple,korinek2026measuring}.

\textbf{Cost and energy efficiency.} \emph{AI Agents
That Matter}, \emph{Cost-of-Pass}, and the \emph{Holistic Agent Leaderboard} compare
performance with financial or usage-based costs
\citep{kapoor2024agentsmatter,erol2025costofpass,kapoor2025hal}. Energy studies such as
\emph{Power Hungry Processing}, \emph{ML.ENERGY}, and \emph{MLPerf Power} directly measure energy
use across models, workloads, and accelerators
\citep{luccioni2024power,chung2025mlenergy,tschand2024mlperfpower,
saadfalcon2025ipw}. More recent
work, including \emph{SWEnergy} and \emph{Sustainable Web Agents}, extends energy measurement
to AI-system workflows \citep{tripathy2026swenergy,krupp2026sustainable}.
NVJ and TSJ extend this resource accounting to net dollar value and time saving
rather than benchmark scores or win--tie rates alone.

\paragraph{From token efficiency to economic return.}
Token-level studies vary batch size, context length, and GPU count, including
phase-specific prefill and decoding measurements
\citep{samsi2023words,niu2026tokenpowerbench}. Longer generations can lower
joules per token by amortizing prefill while increasing total request energy
\citep{vellaisamy2026characterization}; long contexts enlarge attention and
key--value-cache traffic. For a common workload, window, and energy boundary,
tokens per joule multiplied by net value or time saving per output token
gives NVJ or TSJ, after unit conversion. Token efficiency therefore complements
output quality, task value, and time saving; tool execution and waiting also
consume energy without producing model tokens. Comparing these measures requires
matched boundaries because GPU counters and MLPerf Power's external whole-system
metering cover different components \citep{tschand2024mlperfpower}.

\section{Expert References and Annotations}
\label{app:reference-protocol}

The tasks span heterogeneous formats, including documents,
spreadsheets, presentations, code, images, audio, video, and archive-based
deliverables. The frozen task records also preserve occupation, sector, and
O*NET-SOC identifiers; source-file and deliverable inventories; and the
grading rubric. The expert-only reference was produced by an expert without
AI; its recorded time is decomposed into prompt review,
source-file use, task execution, and verification or quality review.
The expert-with-AI reference was produced by an expert using AI. Its records
add the AI provider, model version, interface, source files
used, external tools, full transcripts, and turn-level
expert-active, AI-generation-wait, and total times.
The supplied assisted-reference records identify Claude Sonnet 4.6 and
Claude Opus 4.7 used through Claude Web Chat.
These are recorded expert completion times. The assisted condition includes
the expert's setup, monitoring, review, and correction, as well as AI waits.
The expert-only and expert-with-AI references were annotated by different people.

The pairwise comparison with an expert deliverable assesses relative judged
quality; it does not establish absolute task quality or realized economic
savings. We use the GDPval expert deliverables as meaningful reference points
without assuming they are error-free.

The rubric helps ground the
comparison in the task requirements even when an expert deliverable has flaws.
Expert-only reference, expert-with-AI reference, and
AI-system deliverables are normalized to the same evidence representation.
Candidate position is balanced exactly
across the archived task-condition comparisons.

The expert-with-AI reference is a supplied assisted completion condition;
it is not a claim about every possible form of assisted professional work.
Its completion time can exceed expert-only time on individual tasks.
The expert-only reference is the cleaner
common reference for autonomous systems because neither its
deliverables nor its output dollar values depend on a particular assisting model,
interface, or expert--AI workflow. We therefore retain it as primary; the
expert-with-AI reference is a secondary, workflow-specific
operational comparison. Appendix~\ref{app:reference-decomposition} separates
changes in judged win--tie outcomes from changes in time valuation.

\section{Benchmark Contamination}
\label{app:contamination}

During evaluation, the models received each task and its required source files,
but not the grading rubric or expert deliverable. We checked AI deliverables
for exact matches to expert files and unexplained identical document or
workbook components and found none.

GPT-OSS~120B and GPT-OSS~20B were released
before GDPval became public~\citep{openai2025gptoss,openai2025gdpvalannouncement}.
The public model cards for GLM~5.2, Qwen3.6-27B, and Gemma~4~31B do not
mention GDPval
\citep{glm2026license,qwen2026license,gemma2026license}.
These checks show no direct copying in the evaluated outputs; they do not
establish every model's training-data contents.

\section{Judge and Evidence Pipeline}
\label{app:judge}

The judging pipeline is model-agnostic and can be configured with
alternative language-model judges. Main economic results use the
frozen judge \texttt{models/gemini-3.1-flash-lite} at temperature 0. Its input
contains the task specification, grading rubric, source evidence, and both
identity-masked candidate deliverables. The evidence pipeline uses full sanitized
semantic text and rendered visual PDFs with a 48-page visual ceiling. It
performs safe archive extraction, static code inspection, and media parsing.
All verdict records used in the analysis parsed without error.

\subsection{Judge consistency}
\label{app:judge-consistency}

Table~\ref{tab:judge-consistency} compares Gemini 3.1 Flash-Lite (the main
judge), Claude Haiku 4.5, and GPT-5 mini on the same GLM~5.2 comparisons,
including all four attempts per task.

\begin{table}[!htbp]
\centering\small
\caption{\textbf{The three judges agree on 72--77\% of exact outcomes.} An AI win or tie receives task credit. Exact agreement
retains win/tie/loss; binary win--tie agreement collapses wins and ties into one category. All comparisons use GLM~5.2 outputs from four attempts per task.}
\label{tab:judge-consistency}
\begin{tabular}{lrr}
\toprule
Pair & Exact agreement & Binary win--tie agreement \\
\midrule
Gemini--Haiku & 77.2\% & 79.2\% \\
Gemini--GPT-5 mini & 74.0\% & 76.4\% \\
Haiku--GPT-5 mini & 72.3\% & 72.8\% \\
\bottomrule
\end{tabular}
\end{table}

\paragraph{Repeated-output agreement.}
For GLM~5.2, the first two independently generated outputs for each of 220
tasks produced 440 Gemini judgments. The win--tie versus loss decision matched
across the two outputs for 171 tasks (77.7\%).

\paragraph{Expert review.}
In a blinded expert review of 32 randomly selected GLM~5.2 task comparisons,
experts agreed with Gemini 3.1 Flash-Lite on 23 of 32 win--tie versus loss
decisions (71.9\%). A separate automated review yielded the same agreement
rate. Human-to-human agreement on related pairwise response judgments has
been reported at 64.7\% \citep{wei2025rocketeval}.

\section{Execution and End-to-End Energy Accounting}
\label{app:execution}

The AI-system runs were executed in concurrent batches. IPW telemetry
sampled the metered components at a configured 0.2-second
interval. The end-to-end measurement window began immediately before the
batch's runs launched and ended when the final run completed. For
components exposing cumulative energy counters, end-to-end energy used the
boundary-aligned difference between the end and start readings. Otherwise,
sampled power was integrated over the same wall-clock interval. Measurements
from all accelerators assigned to a run were summed over the execution.

For every evaluated deployment, IPW measured GPU, CPU, and DRAM energy.
Reported joules sum these component-energy readings over the execution
window~\citep{ipw2026repo,ipw2026telemetry}.
Table~\ref{tab:serving-configurations} gives the serving settings.

\begin{table}[!htbp]
\caption{\textbf{All evaluated deployments use vLLM, 20 concurrent requests, a 131,072-token context limit, and a 32,768-token output limit.} Tensor parallelism is eight for eight-accelerator deployments and one for single-accelerator deployments.}
\label{tab:serving-configurations}
\centering
\begingroup
\setlength{\tabcolsep}{3pt}
\renewcommand{\arraystretch}{1.08}
\footnotesize
\begin{tabular*}{\linewidth}{@{\extracolsep{\fill}}lccccc@{}}
\toprule
Model and accelerator & Inference & \shortstack{Tensor\\parallelism} & \shortstack{Concurrent\\requests} & \shortstack{Context limit\\(tokens)} & \shortstack{Output limit\\(tokens)} \\
\midrule
GLM~5.2, 8$\times$H200 & vLLM & TP8 & 20 & 131,072 & 32,768 \\
Qwen3.6-27B, 8$\times$H100 & vLLM & TP8 & 20 & 131,072 & 32,768 \\
Gemma~4~31B, 8$\times$H100 & vLLM & TP8 & 20 & 131,072 & 32,768 \\
GPT-OSS~120B, 8$\times$H100 & vLLM & TP8 & 20 & 131,072 & 32,768 \\
GPT-OSS~20B, 8$\times$H100 & vLLM & TP8 & 20 & 131,072 & 32,768 \\
Qwen3.6-27B, 1$\times$B200 & vLLM & TP1 & 20 & 131,072 & 32,768 \\
Qwen3.6-27B, 1$\times$H200 & vLLM & TP1 & 20 & 131,072 & 32,768 \\
Qwen3.6-27B, 1$\times$H100 & vLLM & TP1 & 20 & 131,072 & 32,768 \\
Qwen3.6-27B, 1$\times$MI350X & vLLM & TP1 & 20 & 131,072 & 32,768 \\
\bottomrule
\end{tabular*}
\endgroup
\end{table}

We chose OpenHands as the primary harness because its compact tool surface, a
terminal and file editor, was sufficient for producing and validating the
heterogeneous deliverables in GDPval. The run ledger records turns, tool
calls, AI-system completion time, outcome, and energy, allowing the same evaluation protocol to
be applied to other harnesses and inference backends. The codebase also includes
portability paths for MiniMax and Apple M4; these
conditions are not pooled with the main results unless the complete evaluation
protocol is matched.

The active-request allocation in Equation~\ref{eq:active-energy} reconciles
to the retained batch totals. This conservation identity applies to a full
batch; it does not imply that concurrent requests draw equal instantaneous
power.

\section{Energy-Attribution Sensitivity}
\label{app:energy-attribution}

A concurrent batch has one measured energy total, while individual requests
overlap in time. Our primary active-request rule
(Equation~\ref{eq:active-energy}) assigns each moment's energy across the
requests then running. An equal-batch-share alternative assigns the same
fraction of the batch total to every request. Both rules conserve measured
batch energy, but they can assign different energy to individual tasks
because requests finish at different times.

We apply both rules to the evaluated GLM~5.2 task pool underlying
Figure~\ref{fig:category-rankings}, holding AI outputs, judge decisions,
expert times, wages, AI-system execution costs, and GDP weights fixed.
Table~\ref{tab:energy-allocation-categories} gives the category comparisons.
Protective Service remains first in win--tie rate per MJ, and Computer \&
Mathematical remains first in NVJ. The TSJ leader changes. Architecture \& Engineering leads
under the primary active-request rule, while Computer \& Mathematical leads
under equal batch shares. Eight NVJ ranks and nine TSJ ranks change across
the 13 categories. Both rules conserve each measured batch total; neither establishes
the energy drawn by an individual concurrent request.

\begin{figure}[!htbp]
    \centering
    \begin{paperfigureframe}[0.96]
    \includegraphics[width=\linewidth]{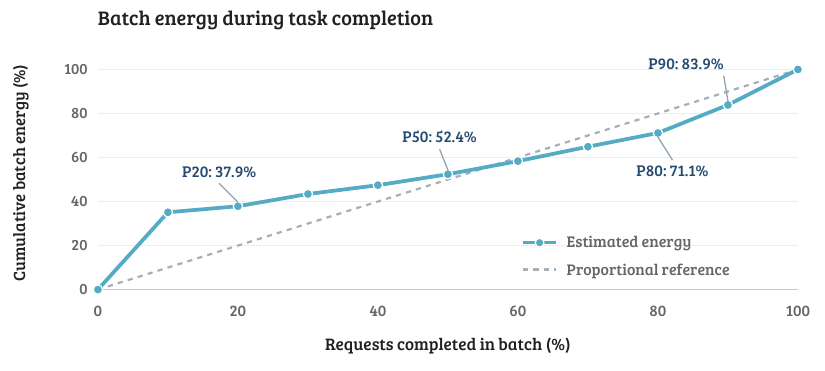}
    \end{paperfigureframe}
    \caption{\textbf{Request-completion share does not track batch energy proportionally.} P20, P50, P80, and P90 mark completion of 20\%, 50\%, 80\%, and 90\% of requests. Each point pools batch totals after estimating energy consumed at that milestone. Intermediate points use each batch's average power and recorded completion times; the 100\% endpoint is measured batch energy. The dotted line shows proportional completion for reference.}
    \label{fig:completion-energy}
\end{figure}

\begin{table}[!htbp]
\centering\small
\caption{\textbf{Equal batch shares change the TSJ leader but leave the NVJ leader unchanged.} Each cell shows active-request allocation / equal batch share. Active-request allocation is the paper's primary method. AI outputs, judgments, expert times, wages, costs, and GDP weights are fixed; only energy assignment changes. Values are point estimates.}
\label{tab:energy-allocation-categories}
\begin{tabular*}{\linewidth}{@{\extracolsep{\fill}}lrrr@{}}
\toprule
Category & Energy (MJ) & NVJ (USD/MJ) & TSJ (hours/MJ) \\
\midrule
Computer \& Mathematical & 0.355 / 0.337 & 523.8 / 551.6 & 7.29 / 7.67 \\
Architecture \& Engineering & 0.495 / 0.597 & 413.9 / 343.2 & 7.89 / 6.54 \\
Management & 0.360 / 0.349 & 374.6 / 387.0 & 6.75 / 6.97 \\
Protective Service & 0.225 / 0.229 & 278.1 / 272.9 & 4.96 / 4.86 \\
Production & 0.550 / 0.786 & 221.1 / 154.6 & 6.12 / 4.28 \\
Community \& Social Service & 0.282 / 0.363 & 188.5 / 146.3 & 6.16 / 4.78 \\
Legal & 0.370 / 0.334 & 149.7 / 166.1 & 1.50 / 1.66 \\
Business \& Financial Operations & 0.505 / 0.481 & 145.6 / 152.9 & 3.32 / 3.49 \\
Personal Care \& Service & 0.401 / 0.503 & 134.3 / 107.2 & 7.15 / 5.70 \\
Arts, Design, Entertainment \& Media & 0.450 / 0.427 & 103.8 / 109.5 & 2.71 / 2.86 \\
Healthcare Practitioners \& Technical & 0.445 / 0.504 & 97.8 / 86.3 & 1.67 / 1.47 \\
Sales \& Related & 0.356 / 0.362 & 65.4 / 64.4 & 2.01 / 1.98 \\
Office \& Administrative Support & 0.604 / 0.567 & 40.1 / 42.6 & 1.84 / 1.96 \\
\bottomrule
\end{tabular*}
\end{table}

Figure~\ref{fig:completion-energy} describes when energy is consumed during
the measured execution batches. For each batch, we find when a given fraction
of its requests has finished and multiply elapsed batch time by its measured
average power. P20, P50, P80, and P90 correspond to an estimated 37.9\%,
52.4\%, 71.1\%, and 83.9\% of pooled batch energy. These intermediate values
use a constant-power approximation; only the full-batch energy total is
directly measured. The figure does not evaluate a stopping policy, since
energy already spent on unfinished requests is included at each milestone
and the outcomes of stopping them are not observed.

\begingroup
\let\section\preprintsection
\section{Calculating AI-System Execution Cost}
\label{app:rental-cost}

We calculate AI-system execution cost for the evaluated runs from recorded
GPU, CPU, and RAM use at published rates. Those rates include hardware and
electricity, so energy is not charged again in dollars.

\paragraph{Rates and resource boundary.}
The eight-H100 comparisons use the July 2026 AWS Linux On-Demand
\texttt{p5.48xlarge} rate of \$55.04 per node-hour. GLM~5.2 uses the
\texttt{p5en.48xlarge} rate of \$63.296 per node-hour for eight H200s.
These are whole-node rates including host CPU and RAM, not rates per GPU.
The calculation also includes one boot volume, one public IPv4 address, and
one final artifact export where applicable
\citep{aws2026pricelist,aws2026p5,aws2026ebs,aws2026vpc,aws2026transfer}.

For the single-NVIDIA-accelerator comparison, we use Modal's published
per-second H100, H200, and B200 prices together with a common 32-core,
256-GiB host reservation \citep{modal2026pricing}. The MI350X comparison
uses CloudRift's published on-demand GPU-hour rate
\citep{cloudrift2026pricing}. These provider prices do not guarantee
identical host specifications or performance across providers.

\paragraph{Time allocation and setup.}
Let \(A_i(t)\) indicate when request \(i\) is active, and let \(N(t)\)
count active requests. We allocate shared active node time using
\(\tau_i=\int A_i(t)/N(t)\,dt\). Shared startup, gaps, and finalization
time \(T_0\) inside a task-execution window contributes
\(T_0\tau_i/\sum_j\tau_j\), so the recorded window is charged once.
Thus the total node time allocated to request \(i\) is
\(T_i^{\mathrm{alloc}}=\tau_i+T_0\tau_i/\sum_j\tau_j\).
The comparison represents a prepared local service; initialization outside
the task-execution window is excluded from the reported execution cost.

\paragraph{Scope.}
Published prices can change. The calculation excludes unobserved
external API charges, durable storage, expert review, development,
coordination, and adoption costs. These belong in a broader deployment
assessment alongside the measured energy and the value of delivered work.

\subsection{Execution-cost sensitivity}
\label{app:provider-cost-sensitivity}
For each evaluated GLM~5.2 task, we calculate the AI-system execution cost by
multiplying its allocated node seconds by the AWS node-hour rate divided by
3,600, then adding its assigned boot-volume and IPv4 charges. To examine a
hypothetical alternative provider price, we replace only that hourly node
rate. For task (i), the resulting cost at rate (r) is
\begin{equation}
C_i(r)=C_i(\mathrm{AWS})+
\frac{T_i^{\mathrm{alloc}}}{3{,}600}(r-63.296),
\end{equation}
where \(T_i^{\mathrm{alloc}}\) includes active and allocated shared node seconds. We use published eight-H200
node tariffs of \$50.44/hour for CoreWeave and \$84.8069/hour for Google Cloud
alongside the \$63.296/hour AWS rate
\citep{coreweave2026pricing,gcp2026a3pricing,aws2026pricelist}.
The same recorded outcomes, time, energy, and non-node charges are held fixed;
the alternative-provider columns are hypothetical price calculations, not
runs on those providers. Table~\ref{tab:provider-cost-counterfactual} shows
the resulting category costs and NVJ. The category NVJ order is unchanged
at these three tariffs, while TSJ is unchanged because execution cost does
not enter its numerator or denominator.

\begin{table}[!htbp]
\centering\footnotesize\setlength{\tabcolsep}{3pt}
\caption{\textbf{Published node tariffs change calculated costs without changing category NVJ order.} For the same recorded GLM~5.2 tasks, only the node-hour price is changed; task outcomes, time, energy, and non-node charges remain fixed. Values are hypothetical costs per category-normalized task and the resulting NVJ, not results of executions on other providers. The AWS column reproduces the evaluated cost boundary.}
\label{tab:provider-cost-counterfactual}
\begin{tabular*}{\linewidth}{@{\extracolsep{\fill}}lrrrrr@{}}
\toprule
& \multicolumn{3}{c}{AI execution cost (USD)} & \multicolumn{2}{c}{NVJ (USD/MJ)} \\
\cmidrule(lr){2-4}\cmidrule(lr){5-6}
Category & CoreWeave & AWS & GCP & CoreWeave & GCP \\
\midrule
Architecture \& Eng. & 1.677 & 2.104 & 2.819 & 414.8 & 412.5 \\
Arts \& Media & 1.582 & 1.986 & 2.660 & 104.7 & 102.3 \\
Business \& Financial & 1.731 & 2.172 & 2.911 & 146.4 & 144.1 \\
Community \& Social & 0.983 & 1.233 & 1.652 & 189.4 & 187.0 \\
Computer \& Math & 1.245 & 1.562 & 2.092 & 524.6 & 522.3 \\
Healthcare & 1.837 & 2.306 & 3.089 & 98.9 & 96.1 \\
Legal & 1.257 & 1.577 & 2.113 & 150.6 & 148.3 \\
Management & 1.256 & 1.576 & 2.112 & 375.5 & 373.1 \\
Office \& Admin. & 2.173 & 2.727 & 3.653 & 41.0 & 38.5 \\
Personal Care & 1.506 & 1.890 & 2.532 & 135.3 & 132.7 \\
Production & 1.859 & 2.333 & 3.125 & 222.0 & 219.7 \\
Protective Service & 0.778 & 0.976 & 1.308 & 279.0 & 276.6 \\
Sales & 1.265 & 1.587 & 2.127 & 66.3 & 63.9 \\
\bottomrule
\end{tabular*}
\end{table}

\endgroup

\section{Economic Weights and Wage Robustness}
\label{app:weights}

GDPval chose sectors for their GDP contribution and sampled predominantly
knowledge-work occupations with large aggregate compensation~\citep{patwardhan2025gdpval}.
Its public gold set has five tasks per occupation, so equal task weights would
reflect the sampling design rather than sector economic scale. We give each of
the nine sectors its normalized GDP share, then divide that share among its
evaluated occupations in proportion to national employment. Sector GDP thus
sets the scale across sectors, while employment distinguishes occupations
within each sector. We do not use wage-bill weights because wages already enter
output dollar value as recorded expert completion time multiplied by the
occupation's matched OEWS median hourly wage~\citep{bls2024oews,patwardhan2025gdpval}.
Using wages again in the weights would give high-wage occupations additional
influence; wage-bill weights would instead emphasize compensation paid to those
occupations. The GDP-based weights preserve sector GDP shares without assuming
equal GDP-to-wage-bill ratios across sectors. The 44 occupations are grouped into 13
occupation categories using the O*NET-SOC taxonomy~\citep{onet2024taxonomy};
O*NET-SOC supplies the classification and does not construct the weights.

Let \(\gamma_s\) be sector \(s\)'s share of U.S. GDP, normalized across the
nine selected sectors, and let
\(L_o=\mathrm{Employment}_o\). The occupation
weight used in both aggregation levels is
\begin{equation}
    w_o^{\mathrm{GDP}}
    =
    \gamma_{s(o)}
    \frac{L_o}{\sum_{j:s(j)=s(o)}L_j},
    \qquad
    \sum_{o\in\mathcal{O}}w_o^{\mathrm{GDP}}=1.
    \label{eq:gdp-weight}
\end{equation}
For completeness, define the per-occupation task average as
\(\widetilde{X}_o=X_o/n_o\), where \(n_o\) is its evaluated task count and
\(X\in\{G,C,S,E\}\), with \(S\) denoting time saving as in
Section~\ref{sec:gdp-aggregation}. For category \(c\), its category weight and within-category
occupation weights are
\begin{equation}
W_c=\sum_{o\in c}w_o^{\mathrm{GDP}},
\qquad
w_{o\mid c}=\frac{w_o^{\mathrm{GDP}}}{W_c},
\qquad
\sum_{o\in c}w_{o\mid c}=1.
\label{eq:within-category-weight}
\end{equation}
The category outcome is
\(\widetilde{X}_c=\sum_{o\in c}w_{o\mid c}\widetilde{X}_o\), while the
GDP-weighted benchmark outcome is
\(\widetilde{X}^{\mathrm{GDP}}=\sum_cW_c\widetilde{X}_c\), with
\(\sum_cW_c=1\). The same within-category and benchmark weights are applied to
output dollar value, AI-system execution cost, time saving and energy before
forming NVJ, TSJ or net value per dollar of AI cost; the method never
averages already-formed occupation or category ratios.

Table~\ref{tab:category-weights} lists the resulting category weights, and
Figure~\ref{fig:economic-weights} visualizes their construction and use.

\begin{figure}[!htbp]
    \centering
    \begin{paperfigureframe}[0.98]
    \includegraphics[width=\linewidth]{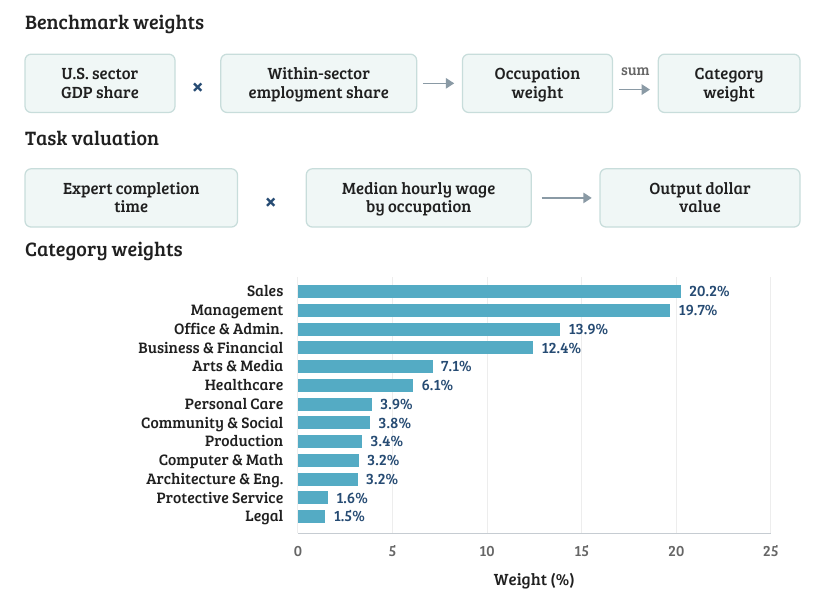}
    \end{paperfigureframe}
    \caption{\textbf{Wages determine output dollar value without also increasing benchmark weights.} Sector GDP shares and within-sector employment shares determine occupation weights. Category comparisons renormalize those weights to one within each category, while the GDP-weighted benchmark metric retains category weights. Expert completion time and occupation-specific median wages determine output dollar value.}
    \label{fig:economic-weights}
\end{figure}

\begin{table}[!htbp]
\caption{\textbf{Sales and Management receive the largest GDP-based category weights.} Employment shares allocate each sector's GDP share among its evaluated occupations; the category weights sum to one and are retained in the GDP-weighted benchmark aggregate.}
\label{tab:category-weights}
\centering
\begingroup
\setlength{\tabcolsep}{3pt}
\renewcommand{\arraystretch}{0.96}
\footnotesize
\begin{adjustbox}{max width=\linewidth,center}
\begin{tabular}{lrrr}
\toprule
Category & Occupations & Weight & Employment (M) \\
\midrule
Management & 7 & 19.68\% & 6.77 \\
Business \& Financial Ops. & 6 & 12.44\% & 3.95 \\
Computer \& Mathematical & 1 & 3.22\% & 1.65 \\
Architecture \& Engineering & 2 & 3.16\% & 0.64 \\
Community \& Social Service & 1 & 3.82\% & 0.38 \\
Legal & 1 & 1.45\% & 0.75 \\
Arts, Design, Entertainment \& Media & 5 & 7.13\% & 0.38 \\
Healthcare Practitioners \& Technical & 3 & 6.10\% & 3.92 \\
Protective Service & 2 & 1.59\% & 0.19 \\
Personal Care \& Service & 2 & 3.91\% & 0.35 \\
Sales \& Related & 8 & 20.24\% & 4.00 \\
Office \& Administrative Support & 5 & 13.87\% & 5.99 \\
Production & 1 & 3.39\% & 0.69 \\
\bottomrule
\end{tabular}
\end{adjustbox}
\endgroup
\end{table}

Table~\ref{tab:wage-robustness} compares mean and median hourly wages in task
valuation while holding GDP-based occupation weights, energy, AI-system execution cost, and
time saving fixed. The headline analysis uses median hourly wages.
U.S. GDP shares, employment, wages, work practices, and estimates of time saving
change over time; results should therefore be recomputed as work changes.
\begin{table}[!htbp]
\caption{\textbf{Mean rather than median wages raise NVJ while leaving TSJ unchanged.} AI-system execution cost is held fixed for GDP-weighted GLM~5.2 with one attempt per task. NVJ is USD/MJ; TSJ is hours/MJ.}
\label{tab:wage-robustness}
\centering
\begingroup
\setlength{\tabcolsep}{3pt}
\renewcommand{\arraystretch}{1.04}
\footnotesize
\begin{adjustbox}{max width=\linewidth,center}
\begin{tabular}{lrrrrr}
\toprule
Wage basis & \shortstack{Output dollar\\value (USD)} & \shortstack{AI-system\\execution cost (USD)} & Energy (MJ) & NVJ & TSJ \\
\midrule
OEWS mean & 85.57 & 1.92 & 0.430 & 194.58 & 3.823 \\
OEWS median & 74.19 & 1.92 & 0.430 & 168.11 & 3.823 \\
\bottomrule
\end{tabular}
\end{adjustbox}
\endgroup
\end{table}

\begingroup
\let\section\preprintsection
\section{Occupation-Category Results}
\label{app:category-results}

Table~\ref{tab:category-success} gives win--tie rates,
Table~\ref{tab:category-unnormalized} gives net dollar value and time saving
per task, and Table~\ref{tab:category-efficiency} gives returns per MJ.
These tables underlie Figure~\ref{fig:category-rankings}. Brackets give 95\% task-composition
intervals. They describe variation across the evaluated tasks and do not
establish a population ordering of categories.

\begingroup
\renewcommand{\arraystretch}{0.94}
\begin{table}[!htbp]
\centering
\footnotesize
\setlength{\tabcolsep}{5pt}
\caption{\textbf{Protective Service leads win--tie rate per MJ, while Personal Care leads win--tie rate.} GLM~5.2, expert-only reference, one attempt per task; occupation outcomes are category-normalized with GDP-based weights. Brackets show 95\% task-composition intervals.}
\label{tab:category-success}
\begin{tabular}{@{}lrr@{}}
\toprule
Category & Win--tie (\%) & Win--tie pp/MJ \\
\midrule
Architecture \& Eng. & 69.00 [38.01, 91.00] & 139.43 [75.84, 229.50] \\
Arts \& Media & 46.09 [28.27, 66.09] & 102.44 [74.15, 136.56] \\
Business \& Financial & 56.81 [37.93, 74.97] & 112.42 [72.81, 168.62] \\
Community \& Social & 60.00 [20.00, 100.00] & 212.94 [60.19, 390.26] \\
Computer \& Math & 80.00 [40.00, 100.00] & 225.08 [122.22, 312.77] \\
Healthcare & 28.23 [8.02, 63.12] & 63.48 [21.77, 107.89] \\
Legal & 60.00 [20.00, 100.00] & 162.14 [49.98, 321.76] \\
Management & 78.67 [62.10, 92.17] & 218.37 [162.96, 289.27] \\
Office \& Admin. & 38.14 [16.10, 62.36] & 63.18 [23.59, 135.67] \\
Personal Care & 95.79 [87.37, 100.00] & 238.63 [150.53, 515.22] \\
Production & 40.00 [0.00, 80.00] & 72.78 [0.00, 150.11] \\
Protective Service & 60.00 [20.80, 99.20] & 267.01 [84.52, 479.38] \\
Sales & 55.31 [34.84, 74.31] & 155.28 [96.17, 222.56] \\
\bottomrule
\end{tabular}
\end{table}

\begin{table}[!htbp]
\centering
\footnotesize
\setlength{\tabcolsep}{4pt}
\caption{\textbf{Architecture \& Engineering leads net dollar value and time saving per task.} GLM~5.2, expert-only reference, one attempt per task; occupation outcomes are category-normalized with GDP-based weights. Brackets show 95\% task-composition intervals.}
\label{tab:category-unnormalized}
\begin{tabular}{@{}lrr@{}}
\toprule
Category & Net dollar value (USD/task) & Time saving (hours/task) \\
\midrule
Architecture \& Eng. & 204.84 [108.93, 288.94] & 3.905 [2.055, 5.569] \\
Arts \& Media & 46.71 [20.18, 79.15] & 1.219 [0.554, 2.028] \\
Business \& Financial & 73.57 [44.46, 106.85] & 1.677 [0.975, 2.508] \\
Community \& Social & 53.12 [12.78, 93.45] & 1.735 [0.423, 3.047] \\
Computer \& Math & 186.16 [68.11, 337.13] & 2.589 [0.924, 4.844] \\
Healthcare & 43.50 [14.21, 86.93] & 0.742 [0.230, 1.541] \\
Legal & 55.40 [7.06, 117.50] & 0.555 [0.046, 1.338] \\
Management & 134.95 [73.68, 229.83] & 2.431 [1.239, 4.337] \\
Office \& Admin. & 24.18 [5.74, 48.62] & 1.111 [0.315, 2.200] \\
Personal Care & 53.92 [34.42, 73.33] & 2.870 [1.753, 3.975] \\
Production & 121.51 [-2.22, 250.73] & 3.361 [0.000, 6.915] \\
Protective Service & 62.49 [7.21, 150.74] & 1.114 [0.128, 2.792] \\
Sales & 23.29 [15.22, 31.58] & 0.717 [0.455, 0.987] \\
\bottomrule
\end{tabular}
\end{table}

\begin{table}[!htbp]
\centering
\footnotesize
\setlength{\tabcolsep}{5pt}
\caption{\textbf{Computer \& Mathematical leads NVJ; Architecture \& Engineering leads TSJ.} GLM~5.2, expert-only reference, one attempt per task; occupation outcomes are category-normalized with GDP-based weights. Brackets show 95\% task-composition intervals.}
\label{tab:category-efficiency}
\begin{tabular}{@{}lrr@{}}
\toprule
Category & NVJ (USD/MJ) & TSJ (hours/MJ) \\
\midrule
Architecture \& Eng. & 413.90 [196.34, 744.61] & 7.890 [3.660, 14.397] \\
Arts \& Media & 103.82 [55.09, 167.63] & 2.710 [1.459, 4.508] \\
Business \& Financial & 145.57 [83.22, 239.58] & 3.318 [1.820, 5.627] \\
Community \& Social & 188.50 [43.10, 382.42] & 6.158 [1.426, 12.522] \\
Computer \& Math & 523.76 [235.93, 770.14] & 7.285 [3.182, 11.080] \\
Healthcare & 97.84 [38.77, 196.72] & 1.669 [0.643, 3.049] \\
Legal & 149.70 [21.88, 370.36] & 1.500 [0.144, 4.257] \\
Management & 374.57 [193.82, 672.56] & 6.748 [3.253, 12.692] \\
Office \& Admin. & 40.05 [8.22, 105.52] & 1.840 [0.470, 4.745] \\
Personal Care & 134.32 [87.59, 261.20] & 7.151 [4.532, 14.139] \\
Production & 221.10 [-4.24, 479.79] & 6.116 [0.000, 13.337] \\
Protective Service & 278.09 [30.55, 669.23] & 4.956 [0.533, 12.268] \\
Sales & 65.40 [42.16, 93.84] & 2.014 [1.283, 2.898] \\
\bottomrule
\end{tabular}
\end{table}

Table~\ref{tab:category-value-time} reports output dollar value and time saving
per win or tie. These components explain why a category can yield fewer wins
and ties per joule while delivering greater dollar value or time saving per
joule.
\begin{table}[!htbp]
\centering
\footnotesize
\setlength{\tabcolsep}{3pt}
\caption{\textbf{Win--tie values and expert times explain why the category rankings differ.} Values condition on outputs judged a win or tie for GLM~5.2 against the expert-only reference. Occupation outcomes are category-normalized with GDP-based weights.}
\label{tab:category-value-time}
\begin{tabular}{@{}lrrrr@{}}
\toprule
Category & Output dollar value (USD) & Expert time (h) & AI time (h) & Time saving (h) \\
\midrule
Architecture \& Eng. & 299.90 & 6.11 & 0.45 & 5.66 \\
Arts \& Media & 105.66 & 3.02 & 0.37 & 2.65 \\
Business \& Financial & 133.31 & 3.34 & 0.39 & 2.95 \\
Community \& Social & 90.58 & 3.22 & 0.32 & 2.89 \\
Computer \& Math & 234.65 & 3.67 & 0.43 & 3.24 \\
Healthcare & 162.29 & 3.04 & 0.41 & 2.63 \\
Legal & 94.96 & 1.31 & 0.38 & 0.93 \\
Management & 173.53 & 3.39 & 0.30 & 3.09 \\
Office \& Admin. & 70.55 & 3.30 & 0.38 & 2.91 \\
Personal Care & 58.26 & 3.40 & 0.40 & 3.00 \\
Production & 309.61 & 9.04 & 0.64 & 8.40 \\
Protective Service & 105.78 & 2.13 & 0.27 & 1.86 \\
Sales & 44.98 & 1.63 & 0.34 & 1.30 \\
\bottomrule
\end{tabular}
\end{table}

\endgroup

\endgroup

\preprintsection{Why Occupation Rankings Differ}
\label{app:success-latency}

Across the evaluated tasks, let
\(a_{i\mid c}=w_{o(i)\mid c}/n_{o(i)}\) be task \(i\)'s normalized
GDP-based weight within category \(c\), and let \(Y_i=1\) when the AI output
wins or ties and zero otherwise. Let \(E_i\) and \(C_i\) denote measured
energy and AI-system execution cost. The category win--tie rate and its
energy-normalized version are
\begin{align}
p_c&=\sum_{i\in c}a_{i\mid c}Y_i,\qquad
\bar E_c=\sum_{i\in c}a_{i\mid c}E_i,\\
\mathrm{WT}_c&=100p_c\ (\%),\qquad
\mathrm{WT/MJ}_c=\frac{100p_c}{\bar E_c}.
\label{eq:success-per-joule}
\end{align}
The latter unit is percentage points per MJ. All attempts contribute energy
and cost, including losses.

For \(p_c>0\), let \(\bar G_{S,c}\) be output dollar value per win or tie and
\(\bar D_{S,c}\) be time saving per win or tie. With
\(\bar C_c=\sum_i a_{i\mid c}C_i\), the two economic returns decompose as
\begin{equation}
\mathrm{NVJ}_c
=\frac{p_c}{\bar E_c}\bar G_{S,c}-\frac{\bar C_c}{\bar E_c},
\qquad
\mathrm{TSJ}_c=\frac{p_c}{\bar E_c}\bar D_{S,c}.
\end{equation}
Thus win--tie frequency per joule, dollar value per win or tie, and time
saving per win or tie play different roles. The category values
are in Tables~\ref{tab:category-success},
\ref{tab:category-efficiency}, and~\ref{tab:category-value-time}.
Table~\ref{tab:category-rank-comparison} shows the resulting rank changes.
Protective Service ranks first on win--tie rate per MJ but fourth on NVJ and
seventh on TSJ. Computer \& Mathematical ranks third on win--tie rate per MJ
but first on NVJ because its dollar value per win or tie is higher. Production ranks
eleventh on win--tie rate per MJ and sixth on TSJ, despite the long expert
completion times of its tested tasks. Architecture \& Engineering leads TSJ.
Each metric therefore answers a
different question about the same attempts.

\begin{table}[!htbp]
\centering
\small
\setlength{\tabcolsep}{5pt}
\caption{\textbf{Different categories lead win--tie rate per MJ, NVJ, and TSJ.} Point-estimate ranks for GLM~5.2 against the expert-only reference, with one attempt per task and category-normalized GDP-based weights; rank 1 is highest.}
\label{tab:category-rank-comparison}
\begin{tabular}{@{}lrrr@{}}
\toprule
Category & Win--tie pp/MJ rank & NVJ rank & TSJ rank \\
\midrule
Architecture \& Eng. & 8 & 2 & 1 \\
Arts \& Media & 10 & 10 & 9 \\
Business \& Financial & 9 & 8 & 8 \\
Community \& Social & 5 & 6 & 5 \\
Computer \& Math & 3 & 1 & 2 \\
Healthcare & 12 & 11 & 12 \\
Legal & 6 & 7 & 13 \\
Management & 4 & 3 & 4 \\
Office \& Admin. & 13 & 13 & 11 \\
Personal Care & 2 & 9 & 3 \\
Production & 11 & 5 & 6 \\
Protective Service & 1 & 4 & 7 \\
Sales & 7 & 12 & 10 \\
\bottomrule
\end{tabular}
\end{table}

\subsection{Full Occupation-Category Results}
\label{app:success-latency-categories}
Table~\ref{tab:category-full-components} gives the
category-normalized credited value, AI execution cost, time saving,
and energy that form each reported ratio. Table~\ref{tab:category-cost-time}
gives net value per dollar of AI-system execution cost and per AI hour for
the same categories. Table~\ref{tab:occupation-full-results} gives the
corresponding point estimates for every evaluated occupation.
GDP-weighting the underlying unrounded components before forming the ratios reproduces the
GLM~5.2 row of Table~\ref{tab:system-cost-time-energy} (37.69 per AI dollar
and 191.3 USD per AI hour).
\begin{table}[!htbp]
\centering\footnotesize\setlength{\tabcolsep}{2pt}
\caption{\textbf{A high win--tie rate alone does not determine credited value or time saving.} Category-normalized GDP-based task averages, expert-only reference. Credited value and time saving are zero on losses; AI-system execution cost and energy include every attempt. Applying $(G-C)/E$ and $S/E$ to the underlying unrounded components reproduces the NVJ and TSJ point estimates in Table~\ref{tab:category-efficiency}.}
\label{tab:category-full-components}
\begin{tabular*}{\linewidth}{@{\extracolsep{\fill}}lrrrrrr@{}}
\toprule
Category & \shortstack{Win--tie\\(\%)} & \shortstack{Credited value $G$\\(USD)} & \shortstack{AI cost $C$\\(USD)} & \shortstack{Time saving $S$\\(hours)} & \shortstack{Energy $E$\\(MJ)} & \shortstack{AI time $L$\\(hours)} \\
\midrule
Architecture \& Eng. & 69.0 & 206.94 & 2.104 & 3.905 & 0.495 & 0.467 \\
Arts \& Media & 46.1 & 48.70 & 1.986 & 1.219 & 0.450 & 0.384 \\
Business \& Financial & 56.8 & 75.74 & 2.172 & 1.677 & 0.505 & 0.443 \\
Community \& Social & 60.0 & 54.35 & 1.233 & 1.735 & 0.282 & 0.321 \\
Computer \& Math & 80.0 & 187.72 & 1.562 & 2.589 & 0.355 & 0.420 \\
Healthcare & 28.2 & 45.81 & 2.306 & 0.742 & 0.445 & 0.299 \\
Legal & 60.0 & 56.97 & 1.577 & 0.555 & 0.370 & 0.368 \\
Management & 78.7 & 136.53 & 1.576 & 2.431 & 0.360 & 0.319 \\
Office \& Admin. & 38.1 & 26.91 & 2.727 & 1.111 & 0.604 & 0.449 \\
Personal Care & 95.8 & 55.81 & 1.890 & 2.870 & 0.401 & 0.393 \\
Production & 40.0 & 123.84 & 2.333 & 3.361 & 0.550 & 0.610 \\
Protective Service & 60.0 & 63.47 & 0.976 & 1.114 & 0.225 & 0.277 \\
Sales & 55.3 & 24.88 & 1.587 & 0.717 & 0.356 & 0.326 \\
\bottomrule
\end{tabular*}
\end{table}

\begin{table}[!htbp]
\centering
\footnotesize
\setlength{\tabcolsep}{5pt}
\caption{\textbf{Computer \& Mathematical leads net value per AI dollar and per AI hour across occupation categories.} GLM~5.2 against the expert-only reference, one attempt per task. Category-normalized GDP-based averages are combined before forming $(G-C)/C$ and $(G-C)/L$, where $G$ is credited output dollar value, $C$ is AI-system execution cost, and $L$ is AI-system completion time in hours. All attempts contribute cost and time. Point estimates use the underlying unrounded components summarized in Table~\ref{tab:category-full-components}; categories are ordered by net value per AI dollar.}
\label{tab:category-cost-time}
\begin{tabular*}{\linewidth}{@{\extracolsep{\fill}}rlrr@{}}
\toprule
Rank & Category & \shortstack{Net value/AI dollar\\(USD/USD)} & \shortstack{Net value/AI hour\\(USD/hour)} \\
\midrule
1 & Computer \& Math & 119.2 & 442.9 \\
2 & Architecture \& Eng. & 97.4 & 438.4 \\
3 & Management & 85.6 & 423.6 \\
4 & Protective Service & 64.0 & 225.8 \\
5 & Production & 52.1 & 199.3 \\
6 & Community \& Social & 43.1 & 165.7 \\
7 & Legal & 35.1 & 150.5 \\
8 & Business \& Financial & 33.9 & 166.1 \\
9 & Personal Care & 28.5 & 137.1 \\
10 & Arts \& Media & 23.5 & 121.8 \\
11 & Healthcare & 18.9 & 145.3 \\
12 & Sales & 14.7 & 71.5 \\
13 & Office \& Admin. & 8.9 & 53.9 \\
\bottomrule
\end{tabular*}
\end{table}

\FloatBarrier
\begingroup
\renewcommand{\baselinestretch}{1}
\footnotesize\setlength{\tabcolsep}{3pt}
\setlength{\LTcapwidth}{\linewidth}
\begin{longtable}{@{}>{\raggedright\arraybackslash}p{.46\linewidth}rrrr@{}}
\caption{\textbf{Occupation-level economic returns vary widely across the evaluated work.} Expert-only reference and one evaluated attempt per task. Win--tie is the fraction of an occupation's evaluated tasks judged a win or tie. Each occupation's task components are averaged before ratios are formed. These are point estimates; all attempts contribute cost and energy.}\label{tab:occupation-full-results}\\
\toprule
Occupation & Tasks & Win--tie (\%) & NVJ (USD/MJ) & TSJ (h/MJ) \\
\midrule
\endfirsthead
\multicolumn{5}{l}{\textit{Table \ref{tab:occupation-full-results} continued}}\\
\toprule
Occupation & Tasks & Win--tie (\%) & NVJ (USD/MJ) & TSJ (h/MJ) \\
\midrule
\endhead
\bottomrule
\endlastfoot
Accountants and Auditors & 5 & 40.0 & 31.5 & 0.53 \\
Administrative Services Managers & 5 & 100.0 & 369.8 & 6.60 \\
Audio and Video Technicians & 5 & 40.0 & 22.7 & 0.53 \\
Buyers and Purchasing Agents & 5 & 40.0 & 249.3 & 6.69 \\
Child, Family, and School Social Workers & 5 & 60.0 & 188.5 & 6.16 \\
Compliance Officers & 5 & 80.0 & 333.9 & 8.45 \\
Computer and Information Systems Managers & 5 & 60.0 & 515.3 & 5.82 \\
Concierges & 5 & 80.0 & 214.2 & 11.18 \\
Counter and Rental Clerks & 5 & 60.0 & 56.4 & 2.71 \\
Customer Service Representatives & 5 & 40.0 & 45.0 & 2.09 \\
Editors & 5 & 80.0 & 407.1 & 10.49 \\
Film and Video Editors & 5 & 60.0 & 116.7 & 2.94 \\
Financial Managers & 5 & 40.0 & 40.9 & 0.33 \\
Financial and Investment Analysts & 5 & 40.0 & 231.6 & 4.35 \\
First-Line Supervisors of Non-Retail Sales Workers & 5 & 20.0 & 82.8 & 1.83 \\
First-Line Supervisors of Office and Administrative Support Workers & 5 & 20.0 & 6.1 & 0.18 \\
First-Line Supervisors of Police and Detectives & 5 & 60.0 & 292.0 & 5.08 \\
First-Line Supervisors of Production and Operating Workers & 5 & 40.0 & 221.1 & 6.12 \\
First-Line Supervisors of Retail Sales Workers & 5 & 60.0 & 65.2 & 2.54 \\
General and Operations Managers & 5 & 80.0 & 1270.8 & 24.74 \\
Industrial Engineers & 5 & 60.0 & 141.8 & 2.44 \\
Lawyers & 5 & 60.0 & 149.7 & 1.50 \\
Mechanical Engineers & 5 & 80.0 & 913.4 & 17.90 \\
Medical Secretaries and Administrative Assistants & 5 & 60.0 & 90.5 & 3.70 \\
Medical and Health Services Managers & 5 & 100.0 & 241.0 & 3.78 \\
News Analysts, Reporters, and Journalists & 5 & 60.0 & 231.4 & 7.59 \\
Nurse Practitioners & 5 & 80.0 & 653.5 & 9.79 \\
Order Clerks & 5 & 20.0 & 16.0 & 0.84 \\
Personal Financial Advisors & 5 & 60.0 & 190.4 & 3.29 \\
Pharmacists & 5 & 60.0 & 162.8 & 2.42 \\
Private Detectives and Investigators & 5 & 60.0 & 93.6 & 3.36 \\
Producers and Directors & 5 & 20.0 & 39.6 & 0.91 \\
Project Management Specialists & 5 & 60.0 & 197.1 & 3.52 \\
Property, Real Estate, and Community Association Managers & 5 & 80.0 & 119.1 & 2.96 \\
Real Estate Brokers & 5 & 40.0 & 26.9 & 0.58 \\
Real Estate Sales Agents & 5 & 40.0 & 28.1 & 0.73 \\
Recreation Workers & 5 & 100.0 & 125.4 & 6.70 \\
Registered Nurses & 5 & 20.0 & 53.4 & 1.05 \\
Sales Managers & 5 & 80.0 & 331.5 & 4.36 \\
Sales Representatives, Wholesale and Manufacturing, Except Technical and Scientific Products & 5 & 60.0 & 72.4 & 1.81 \\
Sales Representatives, Wholesale and Manufacturing, Technical and Scientific Products & 5 & 40.0 & 99.3 & 1.85 \\
Securities, Commodities, and Financial Services Sales Agents & 5 & 100.0 & 154.3 & 3.56 \\
Shipping, Receiving, and Inventory Clerks & 5 & 40.0 & 61.0 & 2.96 \\
Software Developers & 5 & 80.0 & 523.8 & 7.29 \\
\end{longtable}
\endgroup

\subsection{Task-composition sensitivity}
\label{app:category-omission}

The category intervals in
Tables~\ref{tab:category-success} and~\ref{tab:category-efficiency}
come from 10,000 task-composition bootstrap draws. Each draw resamples evaluated
tasks within occupations, then recomputes the weighted dollar-value,
saved-time, cost, and energy components before forming each ratio.
Recorded judgments and resource measurements stay fixed.

We also test category-rank stability by independently removing one uniformly
chosen task from each occupation category in 10,000 portfolios. For each
portfolio, we re-average the retained tasks within their occupations and
recompute win--tie rate per MJ, NVJ, and TSJ using the same GDP occupation
weights. This changes task composition while keeping the recorded judgments,
AI-system execution costs, and energy measurements fixed.
Table~\ref{tab:task-omission} reports the resulting rank correlations and
how often each original leader stays first. Computer \& Mathematical retains
the NVJ lead in 76.5\% of portfolios, while Architecture \& Engineering
retains the TSJ lead in 40.4\%. These frequencies describe the specified task removals, not the
probability that a category would lead across all professional work.

\begin{table}[!htbp]
\centering\small
\caption{\textbf{The NVJ leader survives one-task deletions more often than the TSJ leader.} Across 10,000 independently drawn portfolios, one task is chosen uniformly from each category and the remaining tasks are re-averaged within their occupations. GDP occupation weights and recorded outcomes and resource use stay fixed. Percentile ranges describe these perturbations, not confidence intervals.}
\label{tab:task-omission}
\begin{tabular*}{\linewidth}{@{\extracolsep{\fill}}lrrr@{}}
\toprule
Ranking & Median $\rho$ & 5th--95th $\rho$ & Leader retained \\
\midrule
Win--tie rate per MJ & 0.951 & 0.896--0.989 & 51.0\% \\
NVJ & 0.962 & 0.824--0.995 & 76.5\% \\
TSJ & 0.918 & 0.813--0.973 & 40.4\% \\
\bottomrule
\end{tabular*}
\end{table}

\section{Expert-Reference and Repeated-Run Sensitivity}
\label{app:reference-results}

\subsection{Separating reference judgments from time valuation}
\label{app:reference-decomposition}

The expert reference affects the metrics through its deliverable, which determines
whether the autonomous output wins or ties, and its expert completion time, which determines
the output dollar value and expert time assigned to that outcome. We cross these two choices for the same
GLM~5.2 outputs under the one-run policy, keeping wages, occupation weights,
AI-system completion time, AI-system execution cost, and energy fixed. The
calculation reuses the recorded runs and judge decisions.

Let \(s,t\in\{\mathrm{E},\mathrm{E{+}AI}\}\) index the judgment and time
references separately, with E denoting the expert-only reference and E+AI denoting the
expert-with-AI reference. Let \(Y_i^s\) indicate a win or tie against deliverable
\(s\), \(H_i^t\) be expert completion time, \(W_i\) the median occupation
wage, \(L_i\) the evaluated run's AI-system completion time, and \(C_i,E_i\) its AI-system execution cost and
energy. With normalized task weights \(p_i=w_{o(i)}^{\mathrm{GDP}}/n_{o(i)}\),
where \(n_{o(i)}\) is the number of evaluated tasks in occupation \(o(i)\),
\begin{align}
\mathrm{NVJ}(s,t)
&=\frac{\sum_i p_i Y_i^s H_i^t W_i-\sum_i p_i C_i}{\sum_i p_i E_i},
\label{eq:cross-reference-npj}\\
\mathrm{TSJ}(s,t)
&=\frac{\sum_i p_i Y_i^s(H_i^t-L_i)}{\sum_i p_i E_i}.
\label{eq:cross-reference-tsj}
\end{align}
Negative time saving is retained. All four combinations charge
0.429915~MJ and \$1.91739 in the GDP-weighted task average.
Table~\ref{tab:reference-factorial} reports their returns.

\begin{table}[!htbp]
\centering
\caption{\textbf{Expert-with-AI judgments and completion times each lower reported NVJ and TSJ.}
For GLM~5.2 under \(\mathrm{pass@}1\), the mixed-reference rows separate judgment and valuation effects.
Brackets show percentile 95\% task-composition intervals.}
\label{tab:reference-factorial}
\small
\begin{tabular*}{\linewidth}{@{\extracolsep{\fill}}llrr@{}}
\toprule
Judgment reference & Time reference & NVJ (USD/MJ) & TSJ (hours/MJ)\\
\midrule
Expert-only & Expert-only & 168.11 [126.00, 222.43] & 3.823 [2.870, 5.021] \\
Expert-with-AI & Expert-only & 69.07 [46.04, 95.84] & 1.818 [1.256, 2.486] \\
Expert-only & Expert-with-AI & 84.90 [64.79, 107.49] & 1.990 [1.432, 2.639] \\
Expert-with-AI & Expert-with-AI & 50.04 [30.77, 70.41] & 1.340 [0.814, 1.888] \\
\bottomrule
\end{tabular*}
\end{table}

\paragraph{Paired contrasts and uncertainty.}
We resample evaluated tasks with replacement within each of the 44 occupations
10,000 times (seed 20260726), sharing each draw across all four combinations
and recomputing ratios and percentage reductions. In a separate sensitivity
analysis, we ran each task four times and compared returns using one through
four attempts (Appendix~\ref{sec:passk}). Table~\ref{tab:reference-contrasts}
uses the relevant starting cell as the percentage denominator. The intervals
vary task composition while holding recorded judgments, valuation inputs,
and energy allocation fixed.

\begin{table}[!htbp]
\centering
\caption{\textbf{Changing both judgments and times reduces NVJ by 70.2\% and TSJ by 65.0\%.}
Every change substitutes the expert-with-AI reference for the expert-only reference. Values are percentage
reductions [95\% interval]; a negative endpoint permits an increase.}
\label{tab:reference-contrasts}
\small
\begin{tabular}{p{.45\linewidth}rr}
\toprule
Change and fixed quantity & NVJ reduction (\%) & TSJ reduction (\%)\\
\midrule
Judgments; expert-only reference times fixed & 58.9 [43.0, 72.7] & 52.5 [37.0, 66.5] \\
Times; expert-only reference decisions fixed & 49.5 [33.1, 63.0] & 47.9 [29.1, 63.2] \\
Times; expert-with-AI reference decisions fixed & 27.5 [-1.7, 50.8] & 26.3 [-6.2, 51.8] \\
Judgments; expert-with-AI reference times fixed & 41.1 [20.9, 60.4] & 32.7 [7.2, 54.9] \\
Both judgments and times & 70.2 [54.4, 82.5] & 65.0 [46.3, 79.4] \\
\bottomrule
\end{tabular}
\end{table}

\paragraph{Interaction and attribution.}
The two reductions from the expert-only reference starting point are not additive.
Switching deliverables first gives NVJ \(168.11\to69.07\to50.04\). The first step removes 58.92 percentage points of the original NVJ and the subsequent time substitution removes 11.32 points, totaling 70.23\%. The reverse order gives \(168.11\to84.90\to50.04\). Averaging the two switching orders assigns 56.7\% of the total NVJ decline to reference judgments and 43.3\% to time valuation (TSJ: 53.5\% and 46.5\%).

\paragraph{Judged win--tie outcomes and reference interpretation.}
The AI system wins or ties less often against the expert-with-AI reference
than against the expert-only reference. The assisted deliverables are
therefore harder to match on average under this judge, and the two
references can differ in which outputs win or tie. The judgment contrast
jointly captures reference content and judge behavior.

\subsection{Repeated runs: \texorpdfstring{Pass@\(k\)}{Pass@k} and four-run win--tie thresholds}
\label{sec:passk}

To test how extra attempts affect return, we execute the AI system
independently four times per task. Under
\(\mathrm{pass@}k\), a task earns credit if at least one of \(k\) independent
attempts wins or ties against the matched expert deliverable.
Thus, \(\mathrm{pass@}1\) evaluates one attempt per task, whereas
\(\mathrm{pass@}4\) asks whether any of four attempts wins or ties.
Increasing \(k\) adds both another opportunity to produce a win or tie and that run's
energy to the denominator.

The \emph{four-run win--tie threshold} is a separate analysis. It always
evaluates and charges all four runs, but varies how many of those four must win
or tie. Pass@\(k\) therefore changes the number of opportunities and the energy
counted; the four-run win--tie threshold holds both fixed and changes only the
number of wins or ties required.

\textbf{Additional runs improve win--tie coverage but reduce return per
joule.} Figure~\ref{fig:thresholds} shows \(\mathrm{pass@}1\),
\(\mathrm{pass@}2\), and \(\mathrm{pass@}4\); NVJ and
TSJ decline with \(k\) because the increase in tasks with a win or tie does not offset the
energy of every additional run. These comparisons charge all scheduled attempts,
including attempts after an earlier output wins or ties; they do not evaluate early
stopping or show that every retry policy reduces return per joule. Retry
policy is therefore an efficiency decision as well as a reliability decision.

\begin{figure}[!htbp]
    \centering
    \begin{paperfigureframe}[0.98]
    \includegraphics[width=\linewidth]{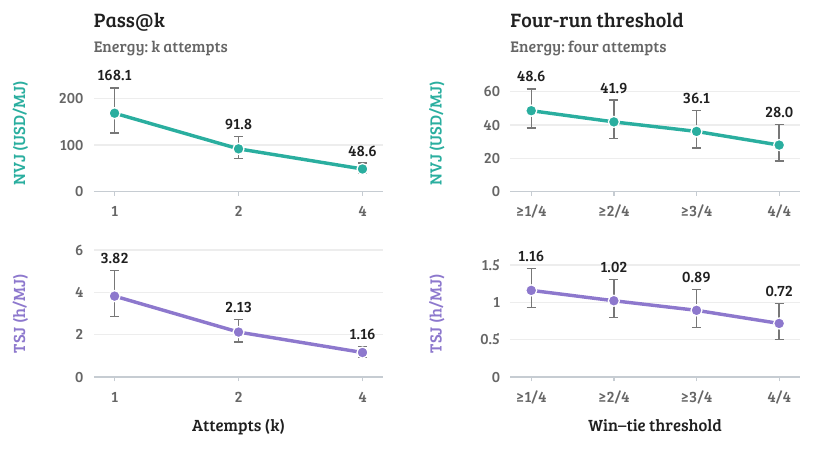}
    \end{paperfigureframe}
    \caption{\textbf{More runs and higher four-run win--tie thresholds both lower measured NVJ and TSJ, but for different reasons.} GLM~5.2 uses the expert-only reference. Left: pass@1, pass@2, and pass@4 include the energy of one, two, and four runs per task. Right: the end-to-end energy of four runs is fixed while the four-run win--tie threshold increases. Whiskers are percentile 95\% task-composition bootstrap intervals.}
    \label{fig:thresholds}
\end{figure}

\textbf{On tasks that match the expert-with-AI reference, the AI system accounts
for most of the measured time saving.} At the \(\geq1/4\) threshold, at least
one of four AI runs wins or ties against the expert-with-AI deliverable. For
those tasks, the AI system saves 1.24 GDP-weighted hours per task relative to
the expert-only reference. Of this total, 0.43 hours (35.0\%) is the difference
between the two expert completion times, and 0.80 hours (65.0\%) is the
difference between expert-with-AI and AI-system completion times.
Appendix~\ref{app:reference-decomposition} separates the effects of changing
the judged deliverable and completion time while holding AI runs and energy fixed.

\begin{figure}[!htbp]
    \centering
    \begin{paperfigureframe}[1]
    \includegraphics[width=\linewidth]{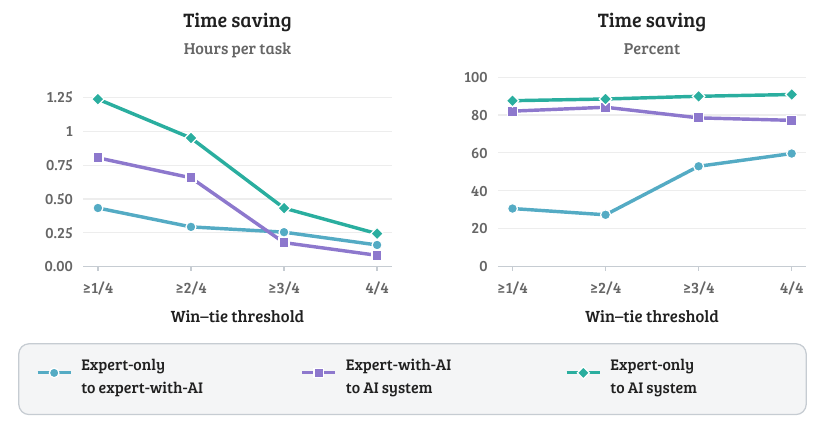}
    \end{paperfigureframe}
    \caption{\textbf{Automation adds time saving beyond expert assistance, but the balance shifts with the success threshold.} At lower thresholds, most saved hours come from automation; at stricter thresholds, more come from expert assistance. GLM~5.2 tasks meet each stated win--tie threshold against the expert-with-AI reference. The left panel reports GDP-weighted hours; the right divides each transition's saved hours by completion time at its starting expert reference. All four runs' energy is retained. Percentages across transitions use different denominators and should not be added.}
    \label{fig:three-way}
\end{figure}

For each task and expert reference, the run order is fixed before evaluating
\(\mathrm{pass@}k\). Let \(Y_{ir}^b\) indicate a win or tie on run \(r\),
\(R_{i,k}^b=\sum_{r=1}^kY_{ir}^b\) count runs that win or tie among the first
\(k\), and \(Q_{i,k}^b=\mathbf{1}(R_{i,k}^b\geq1)\) indicate whether a task
has at least one win or tie. For occupation \(o\), the exact outcome accounting is
\begin{equation}
\begin{aligned}
G_{o,k}^b &= \sum_{i:o(i)=o}Q_{i,k}^bG_i^b, \\
S_{o,k}^b
&= \sum_{\substack{i:o(i)=o\\Q_{i,k}^b=1}}
\left(H_i^b-\frac{\sum_{r=1}^kY_{ir}^bL_{ir}}{R_{i,k}^b}\right), \\
E_{o,k} &= \sum_{i:o(i)=o}\sum_{r=1}^kE_{ir}, \\
C_{o,k} &= \sum_{i:o(i)=o}\sum_{r=1}^kC_{ir}.
\end{aligned}
\label{eq:passk-outcomes}
\end{equation}
Output dollar value is counted at most once. Time saving equals expert completion time
minus mean AI-system completion time among wins and ties when \(Q_{i,k}^b=1\), and zero otherwise; AI-system
completion times for losses are excluded. Energy and AI-system execution cost include all first \(k\) runs without early
stopping, regardless of their judgments. This
best-of-\(k\) convention is also used by
APEX-Agents~\citep{vidgen2026apexagents}.

Table~\ref{tab:human-passk} reports all four \(\mathrm{pass@}k\) policies.

\begin{table}[!htbp]
\caption{\textbf{Four attempts increase output dollar value but reduce NVJ and TSJ.} GLM~5.2 uses the expert-only reference. Every included run incurs cost and energy; output dollar value is counted once. NVJ is USD/MJ and TSJ is hours/MJ. Brackets give conditional 95\% task-composition intervals.}
\label{tab:human-passk}
\centering
\begingroup
\setlength{\tabcolsep}{2pt}
\renewcommand{\arraystretch}{1.04}
\footnotesize
\begin{tabular*}{\linewidth}{@{\extracolsep{\fill}}lrrrrrr@{}}
\toprule
Policy & \shortstack{Output value\\(USD)} & \shortstack{AI cost\\(USD)} & \shortstack{Energy\\(MJ)} & \shortstack{NVJ\\{[95\% CI]}} & \shortstack{TSJ\\{[95\% CI]}} & \shortstack{Net value/\\AI cost} \\
\midrule
\(\mathrm{pass@}1\) & 74.19 & 1.92 & 0.430 & 168.1 [126.0, 222.4] & 3.82 [2.87, 5.02] & 37.69 \\
\(\mathrm{pass@}2\) & 85.75 & 4.00 & 0.891 & 91.8 [71.0, 118.6] & 2.13 [1.65, 2.72] & 20.45 \\
\(\mathrm{pass@}3\) & 89.41 & 5.98 & 1.332 & 62.7 [48.9, 80.0] & 1.46 [1.16, 1.85] & 13.96 \\
\(\mathrm{pass@}4\) & 93.42 & 7.92 & 1.760 & 48.6 [38.3, 61.7] & 1.16 [0.93, 1.45] & 10.79 \\
\bottomrule
\end{tabular*}
\endgroup
\end{table}

Tables~\ref{tab:human-thresholds} and \ref{tab:humanai-thresholds} give complete
GDP-weighted benchmark outcomes across four-run win--tie thresholds.

\begin{table}[!htbp]
\caption{\textbf{Stricter four-run thresholds reduce credited output dollar value and returns at fixed energy.} GLM~5.2 uses the expert-only reference. Every included run incurs cost and energy; output dollar value is counted once. NVJ is USD/MJ and TSJ is hours/MJ. Brackets give conditional 95\% task-composition intervals.}
\label{tab:human-thresholds}
\centering
\begingroup
\setlength{\tabcolsep}{2pt}
\renewcommand{\arraystretch}{1.04}
\footnotesize
\begin{tabular*}{\linewidth}{@{\extracolsep{\fill}}lrrrrrr@{}}
\toprule
Policy & \shortstack{Output value\\(USD)} & \shortstack{AI cost\\(USD)} & \shortstack{Energy\\(MJ)} & \shortstack{NVJ\\{[95\% CI]}} & \shortstack{TSJ\\{[95\% CI]}} & \shortstack{Net value/\\AI cost} \\
\midrule
\(\geq1/4\) & 93.42 & 7.92 & 1.760 & 48.6 [38.3, 61.7] & 1.16 [0.93, 1.45] & 10.79 \\
\(\geq2/4\) & 81.58 & 7.92 & 1.760 & 41.9 [31.7, 54.9] & 1.02 [0.79, 1.31] & 9.30 \\
\(\geq3/4\) & 71.53 & 7.92 & 1.760 & 36.1 [26.3, 48.8] & 0.89 [0.67, 1.17] & 8.03 \\
\(4/4\) & 57.12 & 7.92 & 1.760 & 28.0 [18.4, 40.5] & 0.72 [0.50, 0.99] & 6.21 \\
\bottomrule
\end{tabular*}
\endgroup
\end{table}

Against the expert-with-AI reference, requiring a win or tie on at least
three of four runs brings output dollar value below AI-system execution cost,
so NVJ becomes negative at the two strictest thresholds.

For tasks meeting the same threshold against the expert-with-AI reference, total
expert-only reference-to-AI-system time saving decomposes exactly into
expert-only reference-to-expert-with-AI reference and
expert-with-AI reference-to-AI-system components.

\begin{table}[!htbp]
\caption{\textbf{Stricter four-run thresholds turn NVJ negative against the expert-with-AI reference.} GLM~5.2 incurs cost and energy for every included run; output dollar value is counted once. NVJ is USD/MJ and TSJ is hours/MJ. Brackets give conditional 95\% task-composition intervals.}
\label{tab:humanai-thresholds}
\centering
\begingroup
\setlength{\tabcolsep}{2pt}
\renewcommand{\arraystretch}{1.04}
\footnotesize
\begin{tabular*}{\linewidth}{@{\extracolsep{\fill}}lrrrrrr@{}}
\toprule
Policy & \shortstack{Output value\\(USD)} & \shortstack{AI cost\\(USD)} & \shortstack{Energy\\(MJ)} & \shortstack{NVJ\\{[95\% CI]}} & \shortstack{TSJ\\{[95\% CI]}} & \shortstack{Net value/\\AI cost} \\
\midrule
\(\geq1/4\) & 35.11 & 7.92 & 1.760 & 15.4 [10.5, 20.7] & 0.46 [0.32, 0.61] & 3.43 \\
\(\geq2/4\) & 27.70 & 7.92 & 1.760 & 11.2 [6.1, 16.7] & 0.37 [0.23, 0.52] & 2.50 \\
\(\geq3/4\) & 7.71 & 7.92 & 1.760 & -0.1 [-1.9, 2.1] & 0.10 [0.05, 0.16] & -0.03 \\
\(4/4\) & 3.48 & 7.92 & 1.760 & -2.5 [-3.7, -1.1] & 0.05 [0.01, 0.09] & -0.56 \\
\bottomrule
\end{tabular*}
\endgroup

\end{table}

\begin{table}[!htbp]
\caption{\textbf{Most measured time saving occurs between the assisted expert reference and autonomous AI output.} GDP-weighted task-average decomposition under \(\mathrm{pass@}1\). Bracketed values are percentile 95\% task-composition bootstrap intervals.}
\label{tab:time-decomposition}
\centering
\begingroup
\setlength{\tabcolsep}{4pt}
\renewcommand{\arraystretch}{1.12}
\normalsize
\begin{tabular}{@{}lcccc@{}}
\toprule
Transition & \shortstack{Time saving\\(hours)} & \shortstack{Time saving\\(\%)} & \shortstack{Energy\\(MJ)} & \shortstack{TSJ\\(hours/MJ)} \\
\midrule
Expert-only $\rightarrow$ expert-with-AI & \shortstack{0.21\\{\small [-0.04, 0.45]}} & \shortstack{23.0\\{\small [-5.3, 45.9]}} & -- & -- \\
\addlinespace[2pt]
Expert-with-AI $\rightarrow$ AI system & \shortstack{0.58\\{\small [0.34, 0.83]}} & \shortstack{83.7\\{\small [77.8, 87.1]}} & 0.430 & \shortstack{1.340\\{\small [0.814, 1.888]}} \\
\addlinespace[2pt]
Expert-only $\rightarrow$ AI system & \shortstack{0.78\\{\small [0.55, 1.03]}} & \shortstack{87.4\\{\small [84.4, 89.8]}} & 0.430 & \shortstack{1.818\\{\small [1.256, 2.486]}} \\
\bottomrule
\end{tabular}
\endgroup

\end{table}

Figure~\ref{fig:three-way} shows the corresponding four-run threshold
comparison; Table~\ref{tab:time-decomposition} gives the pass@1 decomposition.
The assisted-expert-to-AI-system component contributes 0.58 hours of the
0.78-hour expert-only-to-AI-system time saving in that comparison.

\preprintsection{Five-Model Comparison}
\label{app:systems}

Table~\ref{tab:system-full} reports GDP-weighted model
outcomes across the same evaluated tasks and task-composition bootstrap intervals.

\begin{figure}[!htbp]
    \centering
    \begin{paperfigureframe}[1]
    \includegraphics[width=\linewidth]{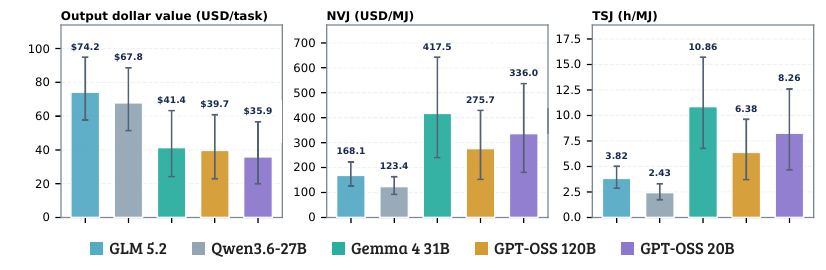}
    \end{paperfigureframe}
    \caption{\textbf{Greater output dollar value need not mean greater economic return per joule.} The five models use GDP-weighted aggregates, the expert-only reference, and one attempt per task. NVJ accounts for execution cost as well as energy; TSJ captures expert time saving per unit of energy. Whiskers are percentile 95\% task-composition bootstrap intervals.}
    \label{fig:global-metrics}
\end{figure}

\begin{table}[!htbp]
\caption{\textbf{GLM~5.2 leads output dollar value; Gemma~4~31B leads NVJ and TSJ among five models.} The models share the evaluated tasks, the expert-only reference, and a one-attempt policy; values are GDP-weighted task averages. NVJ is USD/MJ, TSJ is hours/MJ, and net value per dollar of AI cost divides net dollar value by AI-system execution cost. Brackets give conditional 95\% task-composition intervals. Cost inputs and allocation are in Appendix~\ref{app:rental-cost}.}
\label{tab:system-full}
\centering
\begingroup
\setlength{\tabcolsep}{2pt}
\renewcommand{\arraystretch}{1.04}
\footnotesize
\begin{tabular*}{\linewidth}{@{\extracolsep{\fill}}llrrrrrr@{}}
\toprule
Model & GPUs & \shortstack{Output value\\(USD)} & \shortstack{AI cost\\(USD)} & \shortstack{Energy\\(MJ)} & \shortstack{NVJ\\{[95\% CI]}} & \shortstack{TSJ\\{[95\% CI]}} & \shortstack{Net value/\\AI cost} \\
\midrule
GLM 5.2 & 8$\times$H200 & 74.19 & 1.92 & 0.430 & 168.1 [126.0, 222.4] & 3.82 [2.87, 5.02] & 37.69 \\
Qwen3.6-27B & 8$\times$H100 & 67.81 & 4.12 & 0.516 & 123.4 [92.2, 163.2] & 2.43 [1.73, 3.28] & 15.46 \\
Gemma 4 31B & 8$\times$H100 & 41.38 & 0.86 & 0.097 & 417.5 [239.5, 642.9] & 10.86 [6.76, 15.72] & 47.38 \\
GPT-OSS 120B & 8$\times$H100 & 39.73 & 1.98 & 0.137 & 275.7 [152.5, 429.2] & 6.38 [3.71, 9.63] & 19.09 \\
GPT-OSS 20B & 8$\times$H100 & 35.88 & 1.12 & 0.103 & 336.0 [180.7, 536.4] & 8.26 [4.67, 12.59] & 31.03 \\
\bottomrule
\end{tabular*}
\endgroup
\end{table}

Figure~\ref{fig:global-metrics} summarizes output dollar value, NVJ, and
TSJ, while Table~\ref{tab:system-full} also provides the energy denominator,
GDP-weighted task averages, and intervals.

Occupation-category rankings are broadly similar across models,
especially for output dollar value and time saving. NVJ and TSJ rankings vary
more because the evaluated models have different end-to-end energy profiles. Table~\ref{tab:rank-correlations}
reports all pairwise rank correlations.

\begin{table}[!htbp]
\caption{\textbf{Category output-dollar-value rankings agree across models more than energy-return rankings do on average.} Spearman correlations across the 13 categories after including AI-system execution cost in NVJ.}
\label{tab:rank-correlations}
\centering
\begingroup
\setlength{\tabcolsep}{3pt}
\renewcommand{\arraystretch}{1.04}
\footnotesize
\begin{adjustbox}{max width=\linewidth,center}
\begin{tabular}{llrrrr}
\toprule
Model A & Model B & Output dollar value & Time saving & NVJ & TSJ \\
\midrule
GLM 5.2 & Qwen3.6-27B & 0.830 & 0.703 & 0.780 & 0.544 \\
GLM 5.2 & Gemma 4 31B & 0.654 & 0.912 & 0.560 & 0.731 \\
GLM 5.2 & GPT-OSS 120B & 0.637 & 0.742 & 0.478 & 0.478 \\
GLM 5.2 & GPT-OSS 20B & 0.643 & 0.824 & 0.489 & 0.626 \\
Qwen3.6-27B & Gemma 4 31B & 0.824 & 0.665 & 0.549 & 0.302 \\
Qwen3.6-27B & GPT-OSS 120B & 0.769 & 0.659 & 0.643 & 0.692 \\
Qwen3.6-27B & GPT-OSS 20B & 0.808 & 0.885 & 0.560 & 0.659 \\
Gemma 4 31B & GPT-OSS 120B & 0.802 & 0.808 & 0.863 & 0.654 \\
Gemma 4 31B & GPT-OSS 20B & 0.907 & 0.802 & 0.742 & 0.692 \\
GPT-OSS 120B & GPT-OSS 20B & 0.863 & 0.863 & 0.879 & 0.846 \\
\bottomrule
\end{tabular}
\end{adjustbox}
\endgroup
\end{table}

\FloatBarrier
\preprintsection{Accelerator Comparisons}
\label{app:hardware}

Table~\ref{tab:hardware-full-benchmark} reports the resulting GDP-weighted
outcomes and economic return per joule.

\begin{table}[H]
\caption{\textbf{B200 has the highest NVJ and TSJ among the four single accelerators.} Qwen3.6-27B is evaluated against the expert-only reference under the paper's GDP-based weights. Each accelerator is summarized from its own recorded outcomes and measured energy. Brackets give 95\% task-composition intervals; costs and energy include every attempt.}
\label{tab:hardware-full-benchmark}
\centering
\begingroup
\setlength{\tabcolsep}{3pt}
\renewcommand{\arraystretch}{1.04}
\footnotesize
\begin{adjustbox}{max width=\linewidth,center}
\begin{tabular}{lrr}
\toprule
Accelerator & NVJ (USD/MJ) & TSJ (hours/MJ) \\
\midrule
NVIDIA B200 & 535.5 [388.4, 721.6] & 11.684 [8.416, 15.620] \\
NVIDIA H200 & 391.3 [274.9, 532.4] & 7.293 [4.864, 10.188] \\
NVIDIA H100 & 402.2 [298.6, 531.7] & 7.243 [5.122, 9.934] \\
AMD MI350X & 124.1 [81.7, 177.8] & 1.260 [0.450, 2.297] \\
\bottomrule
\end{tabular}
\end{adjustbox}
\endgroup
\end{table}

One and eight H100s use the same evaluated tasks, expert-only reference,
\(\mathrm{pass@}1\) rule, and GDP weights, with separately recorded outcomes.
Their respective output dollar values, execution costs, and energies are
\((\$64.79,\$0.71,0.159\,\mathrm{MJ})\) and
\((\$67.81,\$4.12,0.516\,\mathrm{MJ})\), yielding NVJ of 402.16 versus
123.39~USD/MJ and TSJ of 7.243 versus 2.432~hours/MJ.

\section{GDPWorks: 83 Additional Occupational Tasks}
\label{app:dataset-extension}

To broaden professional-work evaluation beyond GDPval, we assembled
\emph{GDPWorks}, a dataset of 83 additional tasks created by Snorkel AI. Each task includes
a request, grading rubric, source files, expert-only and expert-with-AI
deliverables, and time annotations. Together with OpenAI's 220 public GDPval
tasks and the annotations added by Snorkel AI, these materials are prepared
for release by the Scaling Intelligence Lab and HazyResearch at
\href{https://huggingface.co/datasets/ScalingIntelligence/joules-to-gdp}{\texttt{ScalingIntelligence/joules-to-gdp}},
with GDPWorks in the \texttt{joules\_to\_gdp} split.

GDPWorks adds 36 detailed O*NET occupations absent from GDPval's 44,
bringing the combined coverage to 80 distinct occupation codes. Examples
include bioinformatics scientists, environmental scientists, and purchasing
managers. The tasks introduce recent work contexts and 2026-dated source
materials, including financial records and environmental monitoring data.
They also target more demanding, source-intensive workflows: median supplied
expert-only time estimates are 4.0 hours, compared with 2.34 hours for GDPval,
and the median number of included reference files is six, compared with one.
These time estimates characterize expected expert effort. GDPWorks
supports future tests of how economic return per joule generalizes to
additional occupations and longer workflows. GDPWorks is not a
GDP-weighted sample; the results reported in this paper use the 220-task
GDPval split.

\FloatBarrier
\endgroup

\end{document}